\documentclass[preprint,sort&compress,11pt,3p]{elsarticle}

\usepackage{moreverb}
\usepackage{mathtools}
\usepackage{amsmath}
\usepackage{algorithmic}
\usepackage{graphics}
\usepackage{graphicx}
\usepackage{caption}
\usepackage{enumitem}
\usepackage{color}
\usepackage{framed}
\usepackage{wrapfig}
\usepackage{bm}
\usepackage{mathrsfs}
\usepackage{mathabx}
\usepackage{multirow}
\usepackage{longtable}
\usepackage{subfigure}
\usepackage{hyperref}
\usepackage{paralist}
\usepackage{indentfirst}
\usepackage{relsize}
\usepackage{extarrows}
\usepackage{lineno,xcolor}
\usepackage{upgreek}
\usepackage{bm}
\usepackage{mwe}
\usepackage{xcolor}
\usepackage{booktabs}
\usepackage[flushleft]{threeparttable}
\usepackage[linesnumbered,lined,ruled]{algorithm2e}

\newcommand{\eref}[1]{(\ref{#1})}

\hypersetup{
bookmarks=true,
bookmarksopen=true,
bookmarksnumbered=true,
unicode=false,
pdftoolbar=true,
pdfmenubar=true,
pdffitwindow=false,
pdfstartview={FitH},
pdftitle={My title},
pdfauthor={Author},
pdfsubject={Subject},
pdfcreator={Creator},
pdfproducer={Producer},
pdfkeywords={keywords},
pdfnewwindow=true,
colorlinks=true,
linkcolor=blue,
citecolor=blue,
filecolor=magenta,
urlcolor=blue,
}

\journal{Journal of Sound and Vibration}

\begin{document}

\title{\LARGE Extracting full-field subpixel structural displacements from videos via deep learning}

\author[NU1]{Lele Luan}
\author[CN]{Jingwei Zheng}
\author[MTU]{Yongchao Yang}
\author[NU1]{Ming L. Wang}
\author[NU1,MIT]{Hao Sun\corref{cor}}
\ead{h.sun@northeastern.edu}

\cortext[cor]{Corresponding author. Tel: +1 617-373-3888}

\address[NU1]{Department of Civil and Environmental Engineering, Northeastern University, Boston, MA 02115, USA}
\address[CN]{China Energy Engineering Corporation Limited, Beijing, 100022, China}
\address[MTU]{Department of Mechanical Engineering, Michigan Technological University, Houghton, MI 49931, USA}
\address[MIT]{Department of Civil and Environmental Engineering, MIT, Cambridge, MA 02139, USA}

\begin{abstract}
	\small
Conventional displacement sensing techniques (e.g., laser, linear variable differential transformer) have been widely used in structural health monitoring in the past two decades. Though these techniques are capable of measuring displacement time histories with high accuracy, distinct shortcoming remains such as point-to-point contact sensing which limits its applicability in real-world problems. Video cameras have been widely used in the past years due to advantages that include low price, agility, high spatial sensing resolution, and non-contact. Compared with target tracking approaches (e.g., digital image correlation, template matching, etc.), the phase-based method is powerful for detecting small subpixel motions without the use of paints or markers on the structure surface. Nevertheless, the complex computational procedure limits its real-time inference capacity. To address this fundamental issue, we develop a deep learning framework based on convolutional neural networks (CNNs) that enable real-time extraction of full-field subpixel structural displacements from videos. In particular, two new CNN architectures are designed and trained on a dataset generated by the phase-based motion extraction method from a single lab-recorded high-speed video of a dynamic structure. As displacement is only reliable in the regions with sufficient texture contrast, the sparsity of motion field induced by the texture mask is considered via the network architecture design and loss function definition. Results show that, with the supervision of full and sparse motion field, the trained network is capable of identifying the pixels with sufficient texture contrast as well as their subpixel motions. The performance of the trained networks is tested on various videos of other structures to extract the full-field motion (e.g., displacement time histories), which indicates that the trained networks have generalizability to accurately extract full-field subtle displacements for pixels with sufficient texture contrast.
\end{abstract}


\maketitle

\section{Introduction}
Displacement measurement is one of the most significant issues for dynamic testing and structural health monitoring (SHM) in infrastructure engineering. Non-contact vibration measurement techniques, such as laser \cite{nassif2005comparison}, radar \cite{zhaocable} and GPS \cite{meng2007detecting}, possess different precision and spatial resolution sensing capacities without the need of sensors installed on the structures. However, these devices are either very costly (e.g., laser sensor) or possess low precision (e.g., GPS). Thanks to recent advances in computer vision techniques, video cameras provide a promising way for sensing of structural vibrations \cite{baqersad2017photogrammetry,feng2018computer,xu2018review,spencer2019advances}. In particular, tracking approaches have been widely used for dynamic displacement extraction (e.g., digital image correlation \cite{hild2006digital,bing2006performance,pan2016digital,wang2018structural,kim2020image} and template matching \cite{fukuda2013vision,feng2015vision,feng2016vision,feng2017experimental,luo2018edge, xiao2020development}). Nevertheless, most of these approaches requires notable features like markers for tracking (e.g., man-made or natural) on the surface of the structures. Furthermore, many approaches fail when the motions are very small such as sub-pixel level movements or when markers are intractable to be installed or selected. On the contrary, optical flow methods can achieve better sub-pixel accuracy by estimating the apparent velocity of movements in the images with remarkable efficiency \cite{fleet1990Computation, gautama2002phase}. 

With the assumption of constant contour of local phase in optical flow, the very recently developed technique of phase-based displacement extraction \cite{chen2015modal,chen2016video} combined with motion magnification \cite{wadhwa2013phase} has significantly promoted the application of video cameras to subpixel motion measurement as well as modal identification and visualization in structural dynamics. Chen \emph{et al.} \cite{chen2015modal} firstly demonstrated this approach for displacement extraction, modal identification and visualization by analyzing a high-speed video recording the vibration of a cantilever beam \cite{chen2015modal}, followed by other applications to real structures such as the antenna on a building \cite{chen2017video} and a steel bridge \cite{wadhwa2017motion,chen2018camera}. The phase-based approach was also applied to measure small motions in other vibration systems like vibropet electrodynamic shaker \cite{diamond2017accuracy} and magnetically system rotor (MSR) \cite{peng2020phase}. Besides, the displacements extracted from phase-based approach and the identified dynamical properties were adopted for damage detection of structures such as wind turbine blade \cite{sarrafi2017mode, poozesh2017feasibility, sarrafi2018vibration, sarrafi2019using} and lab scale structures (e.g., a building model and a cantilever beam) \cite{yang2018reference}. Instead of extracting the full field displacement time histories, Yang \emph{et al.} proposed the phase-based motion representation for output-only modal identification using the family of blind source separation techniques and applied this technique to various bench-scale structures like the cantilever beam, vibrating cable and building model \cite{yang2017blind1,yang2017blind2,yang2019estimation1,yang2020blind}. In addition, Davis \emph{et al.} employed the phase variation to represent small motion features to analyze the recorded objects from videos for sound recovery \cite{davis2014visual}, dynamic video interaction \cite{davis2015image} and material property estimation \cite{davis2015visual}. In spite of the proved effectiveness and promise of the phase-based technique, the complex computational procedure limits its real-time inference capacity, which motivates us to tackle this issue in this study.

In general, full-field displacement extraction belongs to a very popular topic in computer vision which is called optical flow estimation. In two dimensional (2D) optical flow estimation, the 2D displacement field can be determined from apparent motion of brightness patterns between two successive images \cite{horn1981determining}. The optical flow field refers to the displacement vectors for all points in the first image moving to the corresponding locations in the second image. Before the appearance of neural networks, the variational approach was dominant in the computation of optical flow \cite{baker2011database,tu2019survey}. Later, convolutional neural networks (CNNs) were widely leveraged to estimate optical flow. Tu \emph{et al.} \cite{tu2019survey} conducted a detailed survey on CNN-based optical flow methods in three categories: supervised, unsupervised and semi-supervised methods. Initially, optical flow estimation was formulated as an end-to-end supervised learning problem. The first two CNNs for optical flow estimation, FlowNetS and FlowNetC, were proposed by Dosovitskiy \emph{et al.} \cite{dosovitskiy2015flownet} based on an encoder-decoder architecture. Later, by introducing a warping operation and stacking multiple FlowNetS, Ilg \emph{et al.} \cite{ilg2017flownet} proposed FlowNet 2.0 to advance the end-to-end learning of optical flow with improvement on capture of small motions. Ranjan and Black \cite{ranjan2017optical} designed a Spatial Pyramid Network (SPyNet) combining with a classical spatial-pyramid formulation, where large motions are estimated in a coarse-to-fine scheme by warping one image of a pair at each pyramid level based on the current flow estimate and computing a flow update. Hui \emph{et al.} \cite{hui2018liteflownet} presented a lightweight CNN by exploiting an effective structure for pyramidal feature extraction and embracing feature warping rather than image warping used in FlowNet 2.0. Sun \emph{et al.} \cite{sun2018pwc} proposed PWC-Net, a compact but effective CNN model, according to the simple and well-established principles such as pyramidal processing, warping, and cost volume. In addition, other unsupervised  \cite{ren2017unsupervised,wang2018occlusion,yin2018geonet} and semi-supervised \cite{lai2017semi} learning networks were also developed for optical flow estimation.

A key issue for CNN-based approaches lies in the difficulty of accurately obtaining dense (full-field) ground truth labels. In most of the existing studies, the networks were trained with manually synthesized datasets like flying chairs \cite{dosovitskiy2015flownet}, Sintel \cite{butler2012naturalistic}, etc. However, since subtle motions at sub-pixel level between two frames cannot be generated accurately in these training data, the high end-point errors (e.g., more than several pixels) of the trained networks consequently, illustrated in a very recent optical flow research \cite{bar2020scopeflow}, restricts applications to precise displacement extraction. Furthermore, the complexity of realistic photometric effects, e.g., image noise and illumination changes, cannot be reflected in the generated datasets \cite{tu2019survey}. Therefore, leveraging real-world videos has the potential to resolve this challenge. In particular, the phase-based displacement extraction approach discussed previously can extract reliable full-field sub-pixel displacements via computing local amplitudes and local phases, and is promising to serve as a candidate approach for training data generation based on real videos. The objective of this paper is to develop a novel CNN architecture, e.g., sub-pixel flow network (SubFlowNet), for fast and accurate extraction of full-field sub-pixel displacement time histories from videos. The dataset for training the proposed network is generated by processing a real video of a lab-scale vibrating structure using the phase-based approach. The sparsity of the ground truth motion field induced by the texture mask is considered in the network design, further reflected in the loss function definition. The resulting trained network can process the input video in real time, measure accurately subtle displacements for pixels with sufficient texture, and alleviate the tedious computational burden that hinders the phase based-approach from real-time operation. 

The reminder of the paper is organized as follows. Section \ref{CNN} presents the proposed CNN architectures for sub-pixel displacement extraction from videos. Section \ref{phase-based approach} introduces the theory of the phase-based displacement extraction approach and the details of training dataset generation. Section \ref{experimrnt and analysis} discusses the experimental verification results and highlights the generalizability of the trained network. Section \ref{conclusions} draws the conclusions and points out the outlook of future work.

\section{SubFlowNet for full-field sub-pixel displacement extraction} \label{CNN}
To enable fast (e.g., real-time) extraction, we propose a deep learning model based on CNN which maps a pair of video frames to full-field high-resolution displacements at sub-pixel levels. In this section, we discuss the fundamentals of CNN and the proposed network architectures.

\begin{figure}[t!]
	\centering\includegraphics[width=0.7\linewidth]{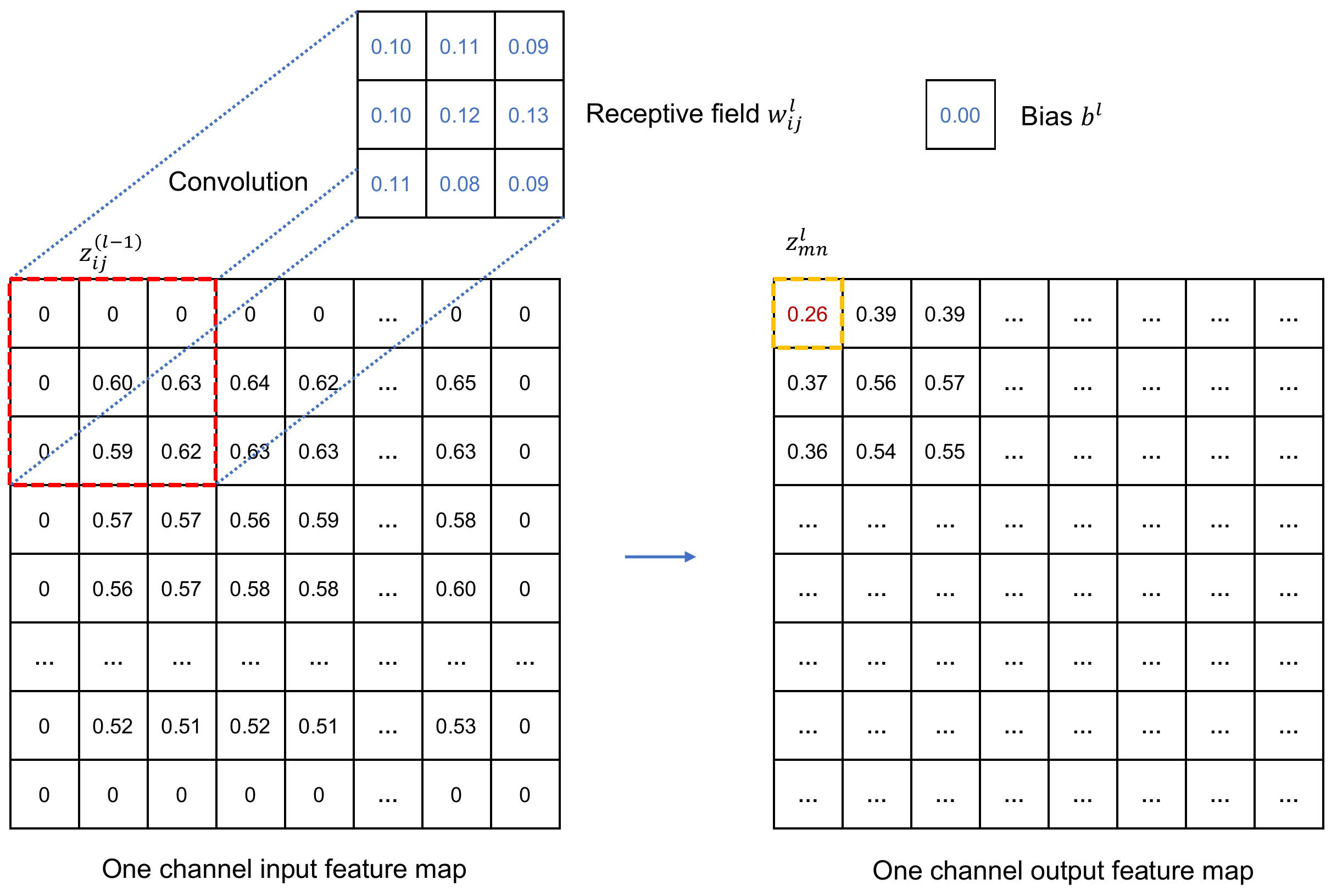}
	\caption{Illustration of the convolution operation. For example, the input is a gray-scale image after blurring and downsampling. Zero-padding is used to keep the resolution of the output as the input. The receptive field (with kennel weights) and bias are learnable parameters. For convenience, the convolution process for only one receptive field generating one-channel output is shown here.}
	\label{fig:convolution_operation}
\end{figure}

\subsection{Fundamentals of convolution neural network}
Let us consider a typical layer of a standard CNN architecture. The basic network unit, convolution (Conv) layer, performs feature learning from the feature map of the input to the current layer. The size of a Conv layer is defined as ``height \(\times\) width \(\times\) depth (input channels) \(\times\) filters (output channels)'', e.g., \(48 \times 48 \times 1 \times 8\). Each Conv layer has a set of learnable kernels (receptive fields) with a size of \(k \times k\) which contains a group of shared weights. The depth of the kernel is determined by the number of channels in the input feature map while the number of kernels determines the number of channels in the output. In the forward propagation, the kernels convolve across the input and compute dot products of the kernel with a local region of the input. Then bias is added to the summation of the dot products producing one single point feature \(z_{mn}^{\left(l\right)}\) in the output. Fig. \ref{fig:convolution_operation} gives an illustration of the convolution operation for one single receptive field. The operation of convolution of the $l$th layer for a single point feature can be written as
\begin{equation}
\label{convolution_operation}
z_{mn}^l=\sum_{i=1}^{3}\sum_{j=1}^{3}{z_{ij}^{\left(l-1\right)}w_{ij}^l}+b^l
\end{equation}
where \(w_{ij}^l\) denotes the receptive field with size of \(3\times3\); \(z_{ij}^{\left(l-1\right)}\) is the \(3\times3\) local region of the input with zero padding; and \(b^l\) is the bias.

\begin{figure}[t!]
	\centering\includegraphics[width=0.7\linewidth]{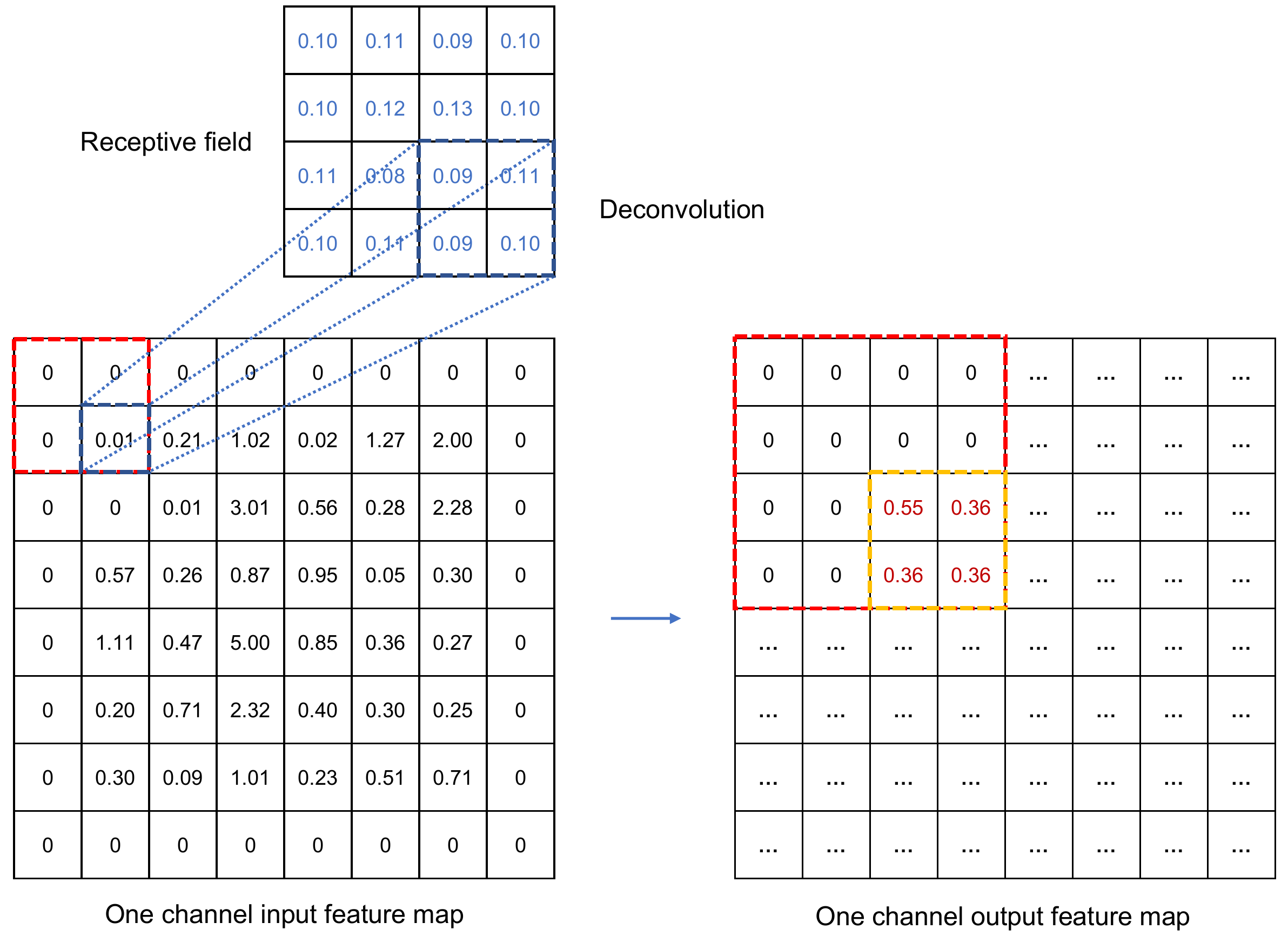}
	\caption{Illustration of the deconvolution operation. The size of receptive field is 4 \(\times\) 4 and the stride is 2. Zero-padding is used to make the resolution of the output twice of the input size. The weights in the receptive field and the responding bias are learnable parameters. The consider example only has one channel for both input and output feature maps.}
	\label{fig:deconvolution_operation}
\end{figure}

In CNN, the pooling operation is often used to reduce the spatial size of the feature map. Hence, the pooling layer is necessary in CNNs to make the network training computationally feasible and, more fundamentally, to allow aggregation of information over large area of the input images \cite{dosovitskiy2015flownet}. Common pooling processes include max pooling and average pooling which take either maximum or mean values from a pooling window. However, these pooling layers may lead to the loss of information of feature maps. Alternatively, convolutions with a stride of 2 are used instead to reduce the spatial size of feature maps as well as to get smaller input feature space. Since these convolution operations result in the reduction of feature map resolution, deconvolution operations (Deconv) are followed to downscale/refine the resolution. Fig. \ref{fig:deconvolution_operation} shows an example of the deconvolution operation. In the Deconv layer, one single point feature in the input feature map is expanded into a 2 \(\times\) 2 matrix and the spatial size of the output is doubled consequentially. Note that no bias is used for the Deconv layers of the designed network architectures in this paper.

The convolution and deconvolution layers are followed by nonlinear activation to introduce nonlinear mapping capability. In this paper, the LeakyReLU function with a slope coefficient 0.1 is employed as the activation function, given by
\begin{equation}
\label{relu_activation}
  f(x) =
  \begin{cases}
  0 & \text{if $x>0$} \\
  0.1x & \text{otherwise}
  \end{cases}
\end{equation}
It should be noted the LeakyReLU activation function is used for all convolution and deconvolution layers except for the final output layer.

\subsection{Design of SubFlowNet architectures}\label{net_ach}
\begin{figure}[t!]
	\centering\includegraphics[width=1.0\linewidth]{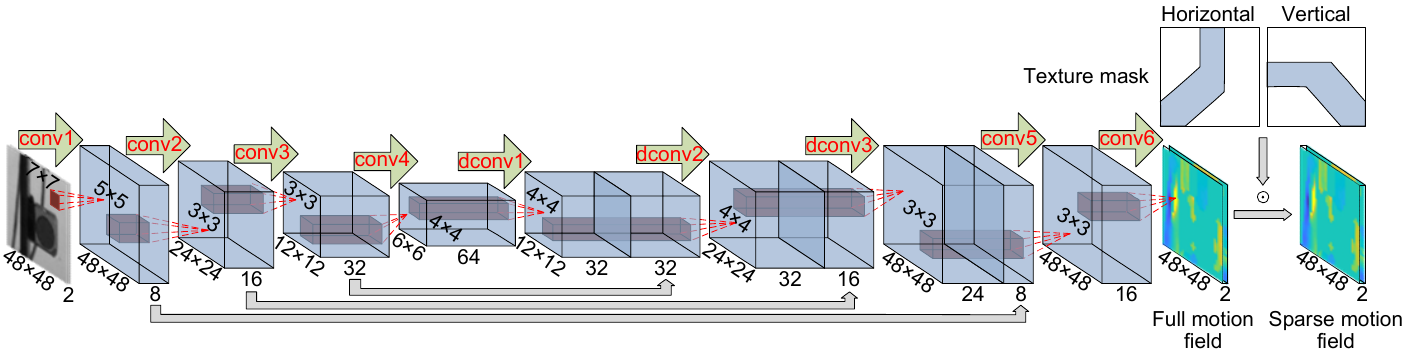}
	\caption{Network architecture of SubFlowNetS, consisting of a series of Conv and Deconv layers. The image pairs (reference frame and current frames) are stacked together and feed into the encoder-decoder architecture for motion representatives extraction. The extracted motion representatives are then mapped to full-field displacement in both horizontal and vertical directions through convolutional layers. The sparse motion field is obtained by applying a texture mask, which imposes hardly the sparse regularization. Training of the network is supervised by both full and sparse motion fields, while the testing of the trained network only keeps the full motion field as the output.}
	\label{fig:network_architecture_s}
\end{figure}

\begin{figure}[t!]
	\centering\includegraphics[width=1.0\linewidth]{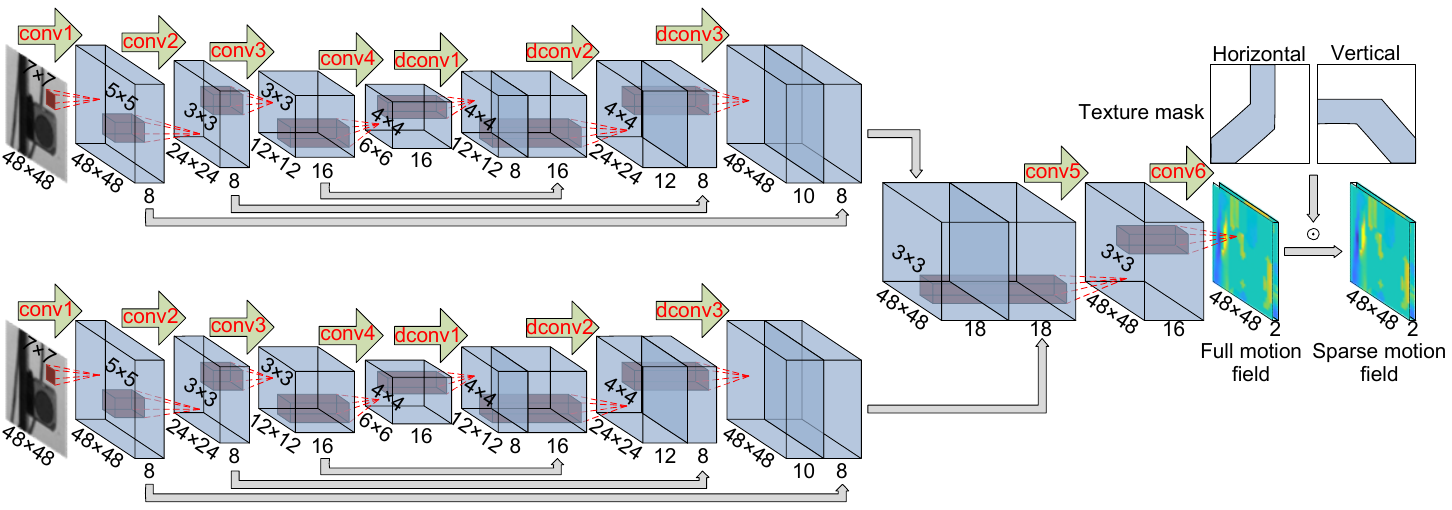}
	\caption{Network architecture of SubFlowNetC. The reference frame and current frames are feed into an identical encoder-decoder branches respectively, consisting of a series of Conv and Deconv layers, to extract the motion representatives. The concatenated motion representatives are then transformed to the full-field displacement in both horizontal and vertical directions through convolutional layers. Other details of the network are same as those of SubFlowNetS in Fig. \ref{fig:network_architecture_s}.}
	\label{fig:network_architecture_c}
\end{figure}

The desired output of the network in this paper is the full-field displacement which has the same resolution as the input. Hence no fully connected layers are employed in this network. Besides, we employ an encode-decoder architecture (with convolution and deconvolution operations) as the basic architecture, which has been proved effective in extracting the features at different scales from the input \cite{badrinarayanan2017segnet}. In particular, the encoder extract lower-dimensional latent features from the input, while the decoder maps the low-dimensional representations back to the original dimension space to form the output. Inspired by the networks developed in \cite{dosovitskiy2015flownet}, we present two CNN architectures (i.e., SubFlowNetS and SubFlowNetC as shown in Figs. \ref{fig:network_architecture_s} and \ref{fig:network_architecture_c}) to extract full-field displacement from video frames. Note that ``Sub'' means that the networks are expected to have the capacity to extract subpixel-level motions. In SubFlowNetS (Fig. \ref{fig:network_architecture_s}), the reference and current frames are stacked together as the input being feed into an encoder-decoder network (with $3\times3$, $4\times4$, $5\times5$ and $7\times7$ kernels), which allows to extract the motion field between the frame pair using a series of Cov/Deconv layers. In SubFlowNetC (Fig. \ref{fig:network_architecture_c}), two separate processing streams, with an identical encoder-decoder architecture, are created for the reference frame and the current frame, respectively, to learn representations, where the feature maps produced from both streams are then combined (concatenated) through transformation layers to form the motion field.

It is noted that the encoder consists of four Conv layers with the kernel sizes of \(7\times7\), \(5\times5\), \(3\times3\) and \(3\times 3\), respectively, while the decoder is composed of three Dcov layers with the kernel sizes of \(4\times4\), \(4\times4\) and \(3\times 3\), respectively. Here, a stride of 2 is used in the Conv layers to reduce the spatial size of feature maps. Another important feature of the encoder-decoder architecture is that the features maps from the encoder are added to those of the decoder, where the features maps exhibit identical resolutions, in a ``residual'' manner to retain the information which may be lost due to the stride in the convolution operations. Since both horizontal and vertical displacement fields are extracted from the image pairs, the two channels in the output represent the motions in these two directions. For both networks as shown in Figs. \ref{fig:network_architecture_s} and \ref{fig:network_architecture_c}, the texture mask is applied to regularize the learning to form sparse motion fields. In particular, the predicted full motion fields are multiplied by a texture mask to produce the sparse motion fields. The networks will be trained based on both labeled motion fields (e.g., full and sparse). The use of extra sparse motion fields can enable the network to pay more attention to the most reliable pixel motions where clear textures are present, leading to more a reliable prediction.

\section{Phase-based approach and dataset generation} \label{phase-based approach}

In order to train the proposed networks, full-field subtle displacements should be provided as labeled training data, which are different from the synthetic datasets used for optical flow estimation neural networks. In particular, the phase-based displacement extraction method \cite{chen2015modal}, which has the capacity to capture sub-pixel displacements, is used to generate the training dataset.

\subsection{Phase-based displacement extraction approach}

\begin{figure}[t!]
	\centering\includegraphics[width=0.45\linewidth]{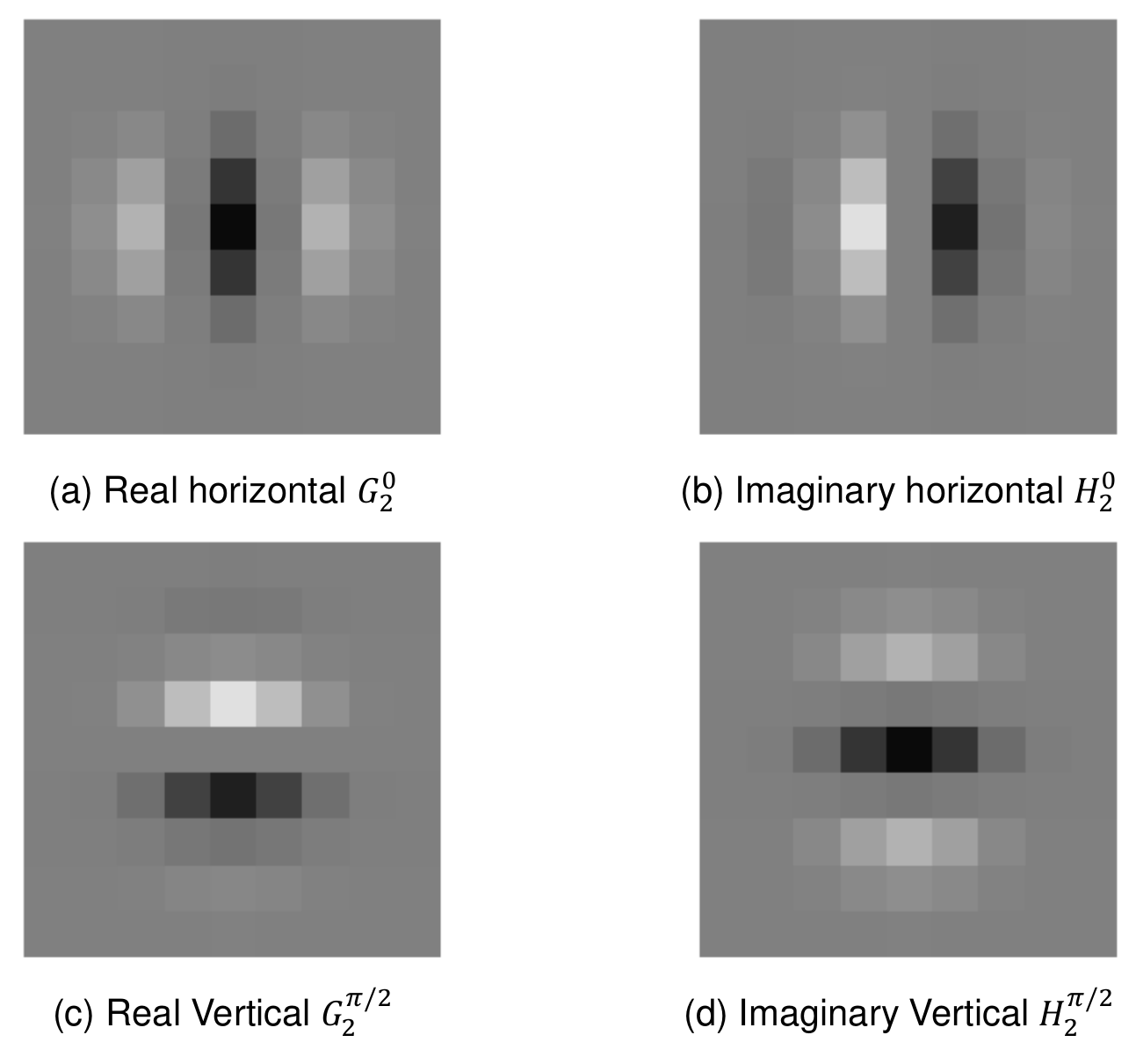}
	\caption{Filters used to compute the local phase and local amplitude. These images represent a 9 by 9 grid of numbers where the gray level corresponds to the value of the filter \cite{chen2015modal}.}
	\label{fig:quadrature_filters}
\end{figure}

In the phase-based approach, the local phase which has been demonstrated to correspond to motion in a video can be obtained with an oriented Gabor filter defined as a sinusoidal wave multiplied by a Gaussian function. Compared with conventional image transform like Fourier transform where the amplitude and phase indicate the global information of the image, the phase-based processing encodes the information of local motion (e.g., local amplitude and local phase). The theory of the phase-based displacement extraction approach is presented as follows. For a given video, with image brightness specified by \(I(x,y,t)\) at a generic spatial location \((x,y)\) and time \(t\), the local phase and local amplitude in orientation \(\theta\) of a frame at time \(t_0\) is computed by spatially bandpassing the frame with a complex filter \( G_\theta^2 + iH_\theta^2 \) to obtain
\begin{equation}
\label{loca_phase}
A_\theta\left(x,y,t_0\right)e^{i\phi_\theta\left(x,y,t_0\right)}=
\left(G_2^\theta+iH_2^\theta\right)\otimes\ I\left(x,y,t_0\right)
\end{equation}
where \(A_\theta(x,y,t_0)\) denotes the local amplitude and \(\phi_\theta(x,y,t_0)\) the local phase. The filters \( G_\theta^2 \) and \(H_\theta^2 \) are convolution kernels for processing the video frame and represent a quadrature pair that differs in phase by \(90^0 \), as shown in Fig. \ref{fig:quadrature_filters}. 

In the phase-based approach, the contour of the local phase is assumed to be constant and its motion through time corresponds to the displacement signal \cite{fleet1990Computation, gautama2002phase}. Hence, displacements can be extracted from the motion of constant contour of local phase in time. This can be expressed as
\begin{equation}
\label{constant_contour}
\phi_\theta(x,y,t_0)=c
\end{equation}
where \(c\) denotes some constant. The displacement signal in a single direction then comes from the distance the local phase contours move between the first frame and the current frame. Differentiating with respect to time of Eq. \eref{constant_contour} yields
\begin{equation}
\label{phase_difference}
\left[\frac{\partial\phi_\theta\left(x,y,t\right)}{\partial x},\frac{\partial\phi_\theta\left(x,y,t\right)}{\partial y},\frac{\partial\phi_\theta\left(x,y,t\right)}{\partial t}\right]\cdot\left(u,v,1\right)=0
\end{equation}
where \(u\) and \(v\) are the velocity in the \(x\) and \(y\) directions respectively. Note that \(\partial\phi_0\left(x,y,t\right)/\partial\ y\approx0\) and \(\partial\phi_{\pi/2}\left(x,y,t\right)/\partial\ x\approx0\). The velocities in units of pixel are then obtained:
\begin{equation}
\label{u_velocity}
u=-\left[\frac{\partial\phi_0\left(x,y,t\right)}{\partial x}\right]^{-1}\frac{\partial\phi_0\left(x,y,t\right)}{\partial t}
\end{equation}
and
\begin{equation}
\label{v_velocity}
v=-\left[\frac{\partial\phi_{\pi/2}\left(x,y,t\right)}{\partial y}\right]^{-1}\frac{\partial\phi_{\pi/2}\left(x,y,t\right)}{\partial t}
\end{equation}
The velocity can be integrated to get the horizontal and vertical displacements at time \(t_{0}\), namely,
\begin{equation}
\label{u_disp}
d_x\left(t_0\right)=-\left[\frac{\partial\phi_0\left(x,y,t_0\right)}{\partial x}\right]^{-1}\left[\phi_0\left(x,y,t_0\right)-\phi_0\left(x,y,0\right)\right]\end{equation}
and
\begin{equation}
\label{v_disp}
d_y\left(t_0\right)=-\left[\frac{\partial\phi_{\pi/2}\left(x,y,t_0\right)}{\partial y}\right]^{-1}\left[\phi_{\pi/2}\left(x,y,t_0\right)-\phi_{\pi/2}\left(x,y,0\right)\right]\end{equation}
Noteworthy, displacements in regions with sufficient texture are treated reliable, which are extracted \cite{chen2016video}. To ensure the reliability of the extracted displacement, the displacements shown in Eq. \eref{u_disp} and Eq. \eref{v_disp} are multiplied by the horizontal and vertical texture masks, respectively, which represent the pixels with sufficient texture contrast identified from local amplitudes. As discussed previously, the sparsity of motion field induced by the mask texture is considered in the network architecture design and loss function definition (see Section \ref{net_ach}).

\subsection{Training dataset generation}

\begin{figure}[b!]
	\centering\includegraphics[width=0.85\linewidth]{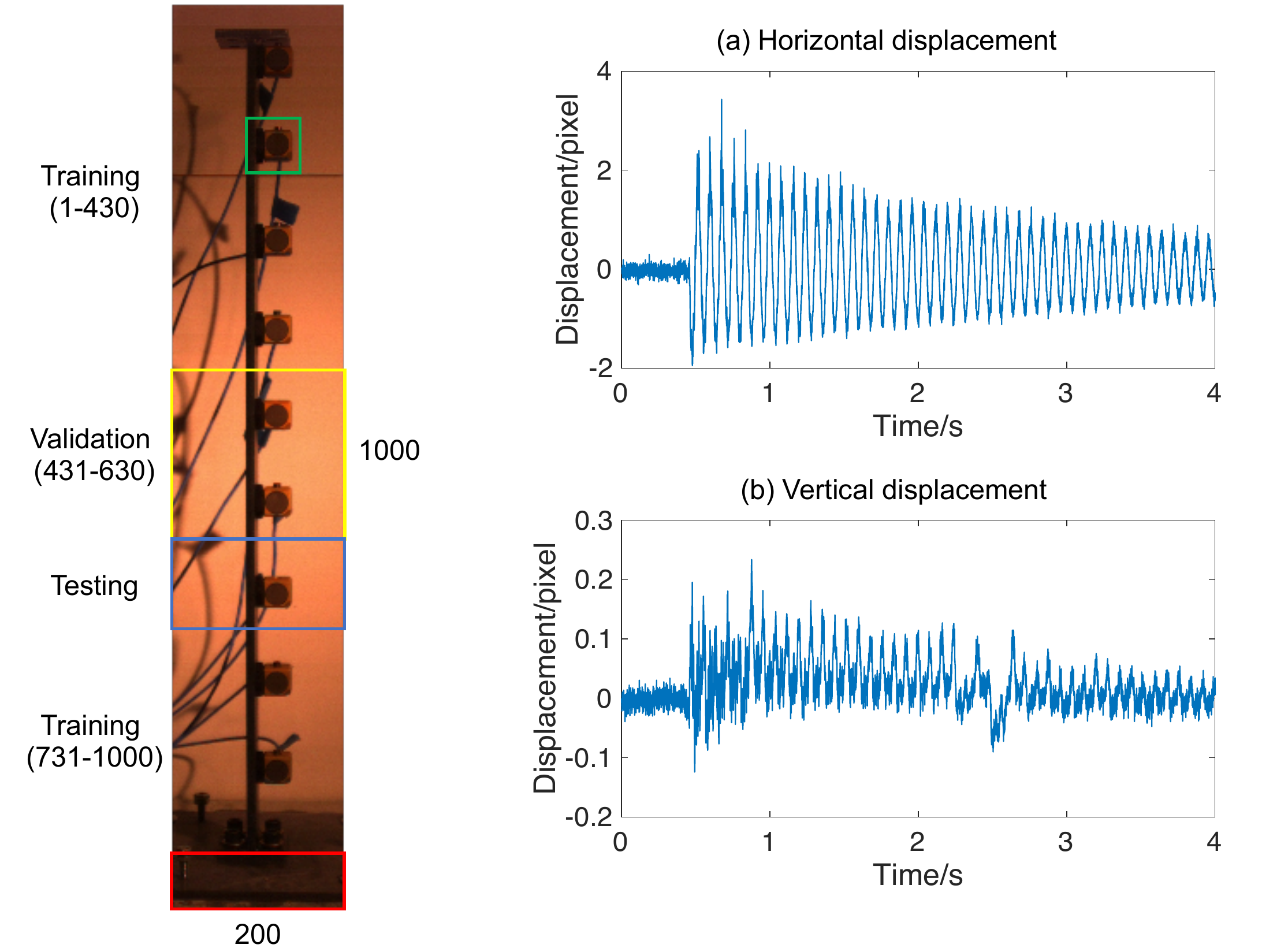}
	\caption{The left figure shows the first frame of the recorded video. The bottom fixed end (in the red box) is excluded in the analysis. The rest of the video is then divided into three segments in the spatial dimension to generate datasets for training, validation and testing. The annotated numbers represent the ranges of these segments along the cantilever. The right figures show the extracted horizontal and vertical time history displacement of a typical pixel at the 2nd accelerometer in the green box.}
	\label{fig:souce_video}
\end{figure}

The training dataset was generated by extracting the displacements of a real video recorded in a lab experiment which \cite{chen2015modal}. The test cantilever beam was excited by a hammer impact near the base. The subsequent vibration was measured by a high-speed camera with the resolution of the video was \(1,056\times200\) at the frame rate of 1,500. Note that the measurements by the accelerometers are not used. The duration of this video is 10 seconds with 15,000 frames in total. For example, Fig. \ref{fig:souce_video} shows the first frame of the high-speed video. The sub-pixel level displacement time history in two directions at the second accelerometer marked with green box extracted by the phase-based approach is also given in Fig. \ref{fig:souce_video}. 

\begin{figure}[t!]
	\centering\includegraphics[width=0.9\linewidth]{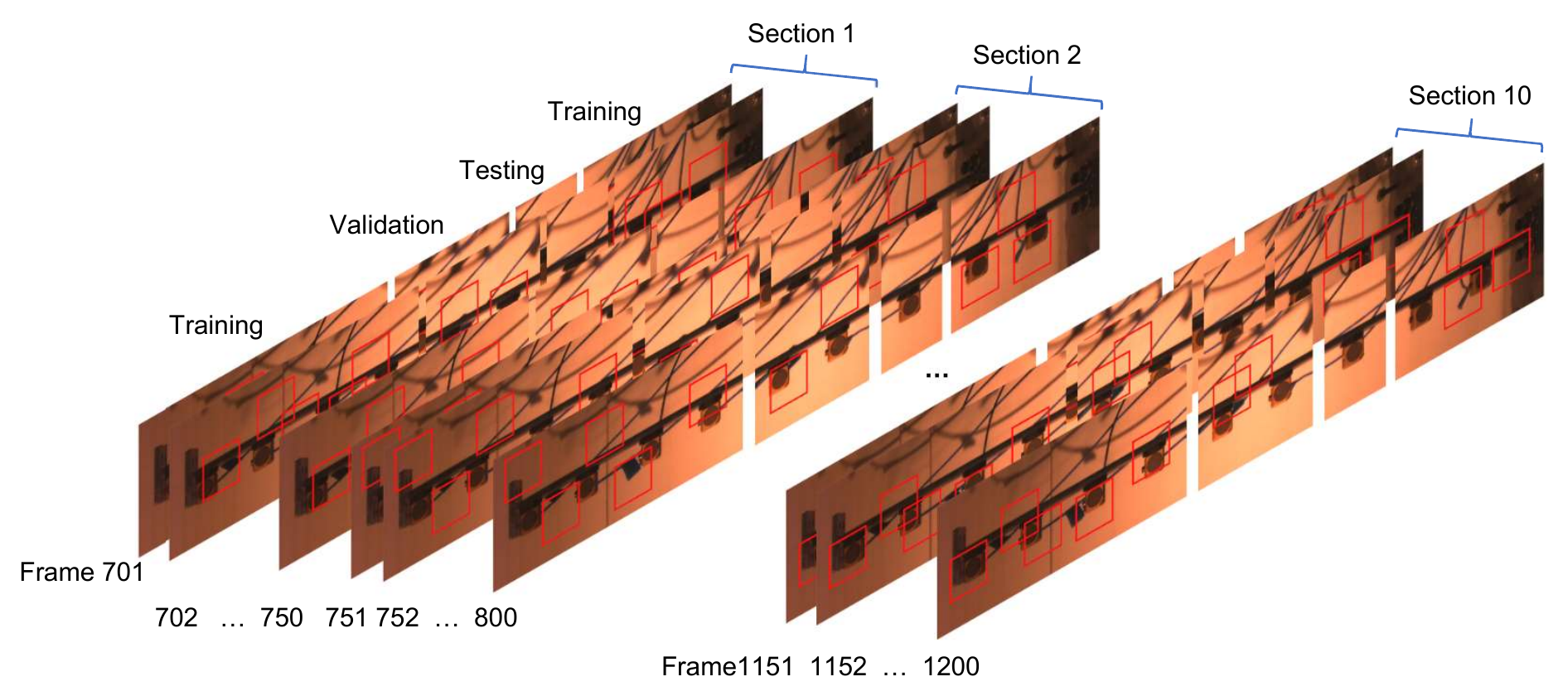}
	\caption{Video segmentation for dataset generation, which is conducted on the original video from frame 701 to 1,200 with the fixed end excluded. In the temporal dimension, the selected video is divided into 10 sections with 50 frames for each.}
	\label{fig:video_segmentation}
\end{figure}

\begin{figure}[t!]
	\centering\includegraphics[width=0.6\linewidth]{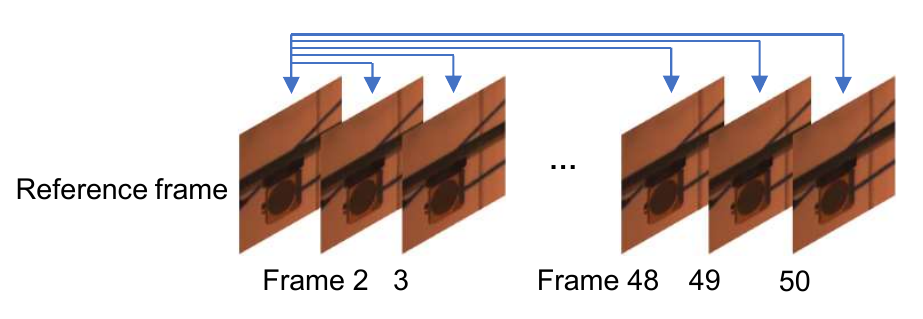}
	\caption{Image pairs in the cropped sub-videos. For each cropped sub-video, the first frame and the subsequent frame are combined to generate an image pair.}
	\label{fig:image_pairs}
\end{figure}

In the optical flow estimation, the full-field displacement can be calculated between two consecutive frames of the video. The two frames are termed as an image pair that serves as input to the proposed network, while the displacement field between these two images is treated as the output. The dataset generation conduced on this high-speed video is introduced as follows. Firstly, the bottom fixed end (in the red box) of the video is excluded, resulting in \(1000\times200\) pixels left. Since the original video mainly has horizontal vibration, a flipped video is implemented to guarantee the generated dataset has both horizontal and vertical displacement samples. In the temporal dimension, 500 frames from frame 701 to 1,200 are selected from the original and flipped videos for dataset generation. Hence, the size of the selected data is \(1,000\times 200 \times500\) in the original video and \(200\times 1,000 \times500\) in the flipped video. The selected parts from both the original and flipped videos are divided into three segments to generate the datasets for training, validation and testing. As shown in Fig. \ref{fig:souce_video}, the area in the yellow box represents validation, the blue box for testing and the rest for training. Each of these three segments is divided into 10 consecutive sections in the temporal dimension as shown in Fig. \ref{fig:video_segmentation}. In each section, the area for training and validation is respectively cropped with 100 and 30 boxes whose size is \(96\times96\) randomly sampled, which produces 100 sub-videos for training and 30 sub-videos for validation. It should be noted that, in order to increase the diversity of the dataset, different sections are cropped with different random boxes. In the cropped sub-videos, the first frame and the subsequent frames are combined as image pairs shown in Fig. \ref{fig:image_pairs} which are fed into the proposed networks as input. As a result, the generated dataset has 100,000 image pairs for training and 30,000 image pairs for validation.

\begin{figure}[t!]
	\centering\includegraphics[width=0.7\linewidth]{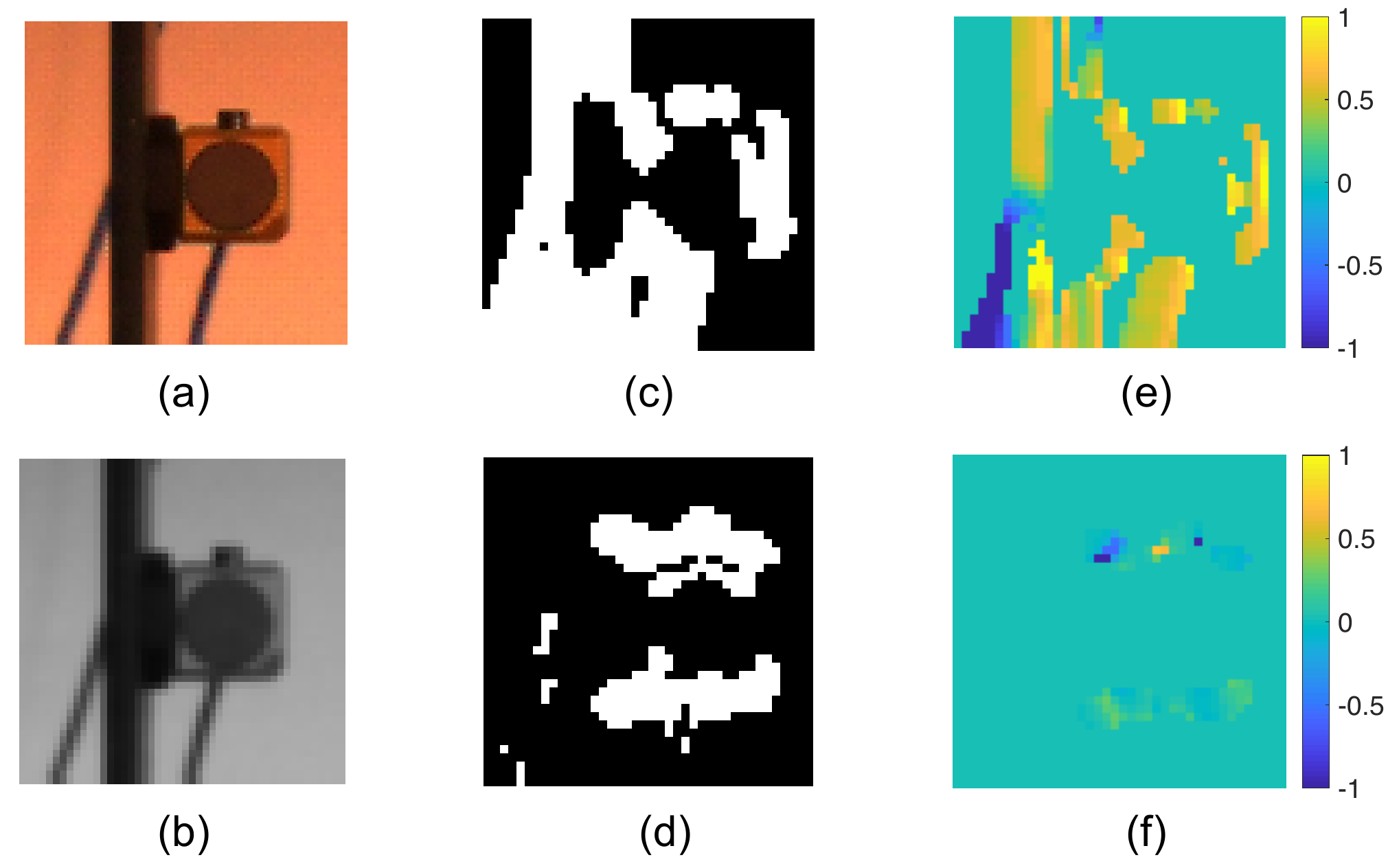}
	\caption{Full-field displacement generation by the phase-based approach for a typical image pair. (a) shows an original study frame before downsampling. (b) shows the frame after blurring and downsampling. (c) and (d) show the masks considering texture and zero-crossing in both horizontal and vertical directions. (e) and (f) are the extracted horizontal and vertical displacement fields.}
	\label{fig:dataset_generation_details}
\end{figure}

After getting the image pairs, the ground truth of the motion field is obtained by the phase-based displacement extraction approach. Firstly, the images in the RGB color space are transformed into the YIQ color space while only Y channel is adopted as the input. Then the images are blurred and downsampled to \(48\times48\) from \(96\times96\), which helps smooth the images and reduce the effect of noise consequently. For example, Fig. \ref{fig:dataset_generation_details}(a) and (b) show an original video frame and the frame after blurring and downsampling. Since displacements are only valid in regions with sufficient texture, displacements in horizontal and vertical directions are extracted separately considering texture masks in these two directions respectively. The texture mask is generated based on the amplitude signal of the first frame after applying the quadrature filter pair. The threshold here is chosen 1/5 of the mean of the 30 pixels with large amplitudes, above which pixels are included in the motion field. Besides, pixels with zero crossing in the \(\partial\phi_0/\partial x\) signal are removed because they cause the displacement calculation to blow up when dividing by a small number close to zero. Fig. \ref{fig:dataset_generation_details}(c) and (d) show the masks in the horizontal and vertical directions respectively considering texture and zero crossing for a typical frame. The extracted displacement fields in both horizontal and vertical directions are shown in Fig. \ref{fig:dataset_generation_details}(e) and (f). It is noted that, although the phase-based displacement extraction method fails to accurately get the displacement in the edges, the resolution of the extracted displacement field is kept the same with the image by using padding in convolution with complex filters, which guarantee the correspondence between the image pixels and their motions.

\section{Experiments and Analysis} \label{experimrnt and analysis}

\subsection{Network and Training Details}

\begin{figure}[t!]
	\centering\includegraphics[width=0.56\linewidth]{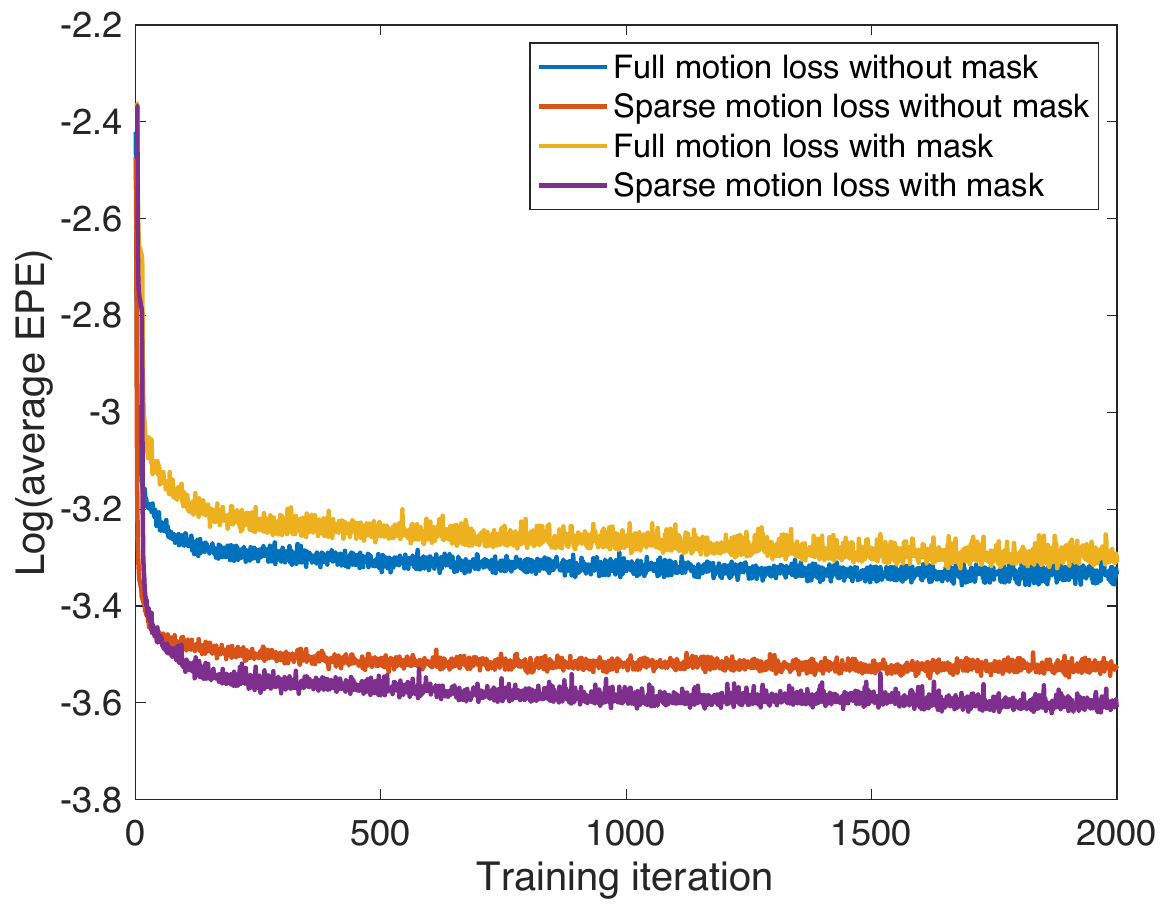}
	\caption{Validation loss comparison on the SubFlowNetC without and with the mask layer. It shows that the mask layer improves the validation accuracy for the pixels with texture mask.}
	\label{fig:validation_loss_comparison}
\end{figure}

The SubFlowNetS and SubFlowNetC networks shown in Fig. \ref{fig:network_architecture_s} and Fig. \ref{fig:network_architecture_c} are trained based on the generated database. In both networks, the deconvolutional kernel size is \(4\times4\). Batch normalization is not used after all Conv and Deconv layers followed by LeakyRelu activation functions. Padding is applied in the convolutions to keep the resolution. The number of trainable parameters (i.e., weights and biases) is 116,024 in SubFlowNetS and 20,840 in SubFlowNetC. The batch size is 128. The loss function is defined as the aggregated end-point error (EPE), the Euclidean distance between the ground truth and the estimated displacement vectors, namely
\begin{equation}
\label{loss_function}
EPE = \underbrace{\frac{1}{N}\sum_{i=1}^{N} {\left\|v_i^{gt}-v_i^{est} \right\|}_2}_{\text{full}} + \underbrace{\frac{1}{N}\sum_{i=1}^{M} {\left\| v_i^{gt}-v_i^{est} \right\|}_2}_{\text{sparse}}
\end{equation}
where \(v_i^{gt}\) is the ground truth displacement vector and \(v_i^{est}\) the estimated displacement vector, for one single pixel \(i\); \(N\) is the number of pixels in the input frames and \(M\) is the number of pixels within the texture mask; \(\|\cdot\|_2\) denotes the \(\ell_2\) norm. Note that EPE has been widely used as loss function in deep learning for optical flow estimation. However, since the phase-based approach only produces accurately the displacements of pixels with masks of sufficient texture contrast, the displacements of the rest of pixels without masks are treated as 0. In order to make the proposed networks learn representations in the regions with sufficient texture contrast, both full and sparse motion fields are employed to supervise the network training. Thus, the total loss function is defined as the summation of the average EPEs on full and sparse ground truth motion fields as shown in Eq. \eref{loss_function}. Although the networks are trained with mask operation, the inference (prediction) by the trained networks only produces the full motion field as output. The Adam optimizer is employed to train the networks given its superiority over other stochastic optimization methods \cite{kingma2014adam}. The learning rate is kept as 0.001 and the total number of training epochs is set to be 2,000. The validation dataset here helps monitor overfitting during training. The trained networks with the lowest validation loss is saved for inference (prediction/testing). To demonstrate the effectiveness of the mask layer, the comparison of validation loss functions is shown in Fig. \ref{fig:validation_loss_comparison}. It is seen that the mask layer clearly improves the validation accuracy the pixels with texture mask.

\subsection{Results}

\begin{figure}[t!]
	\centering\includegraphics[width=0.5\linewidth]{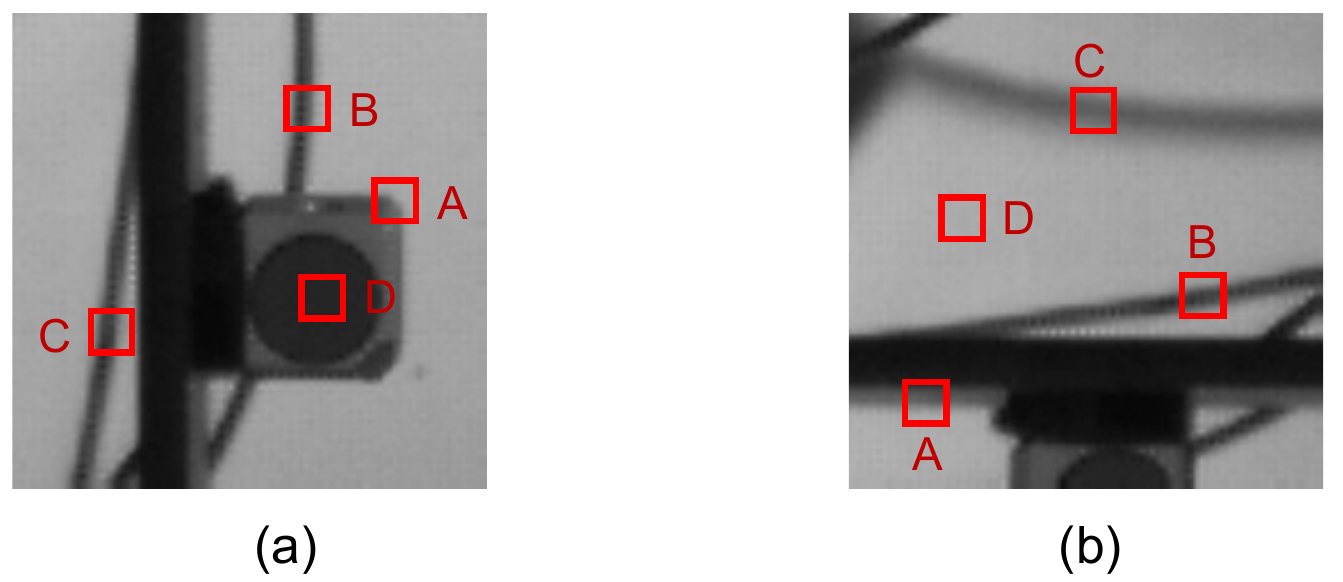}
	\caption{The source testing videos. (a) and (b) show the testing videos cropped from the testing segments in the original and flipped videos in gray scale. The annotated points show the positions of the studied pixels for displacement time history prediction. In the videos, points A, B and C have masks while point D has no mask due to insufficient texture contrast or no motion target.}
	\label{fig:testing_videos_source}
\end{figure}

The training process shows that, for the same dataset, the proposed two networks have similar performance in regard to the validation accuracy (SubFlowNetS produces slightly better accuracy compared to SubFlowNetC). However, due to its larger size of trainable parameters, training SubFlowNetS is much more computational demanding. Hence, the trained SubFlowNetC network is selected for performance demonstration in the rest of the paper. Firstly, the network performance is evaluated on the testing part of the source video as shown in Fig. \ref{fig:souce_video}. The testing videos are generated by cropping the testing segment randomly in both the original and flipped videos. In the temporal direction, 6,000 consecutive frames of the cropped videos are employed for testing. The first frame of the testing video is kept as reference frame and the subsequent ones as study frames. The displacement field of the study frames are estimated while the time histories of selected pixels are used to evaluate the performance of the trained network. Fig. \ref{fig:testing_videos_source}(a) and (b) show typical frames of the testing videos.

\begin{figure}[t!]
	\centering\includegraphics[width=1.0\linewidth]{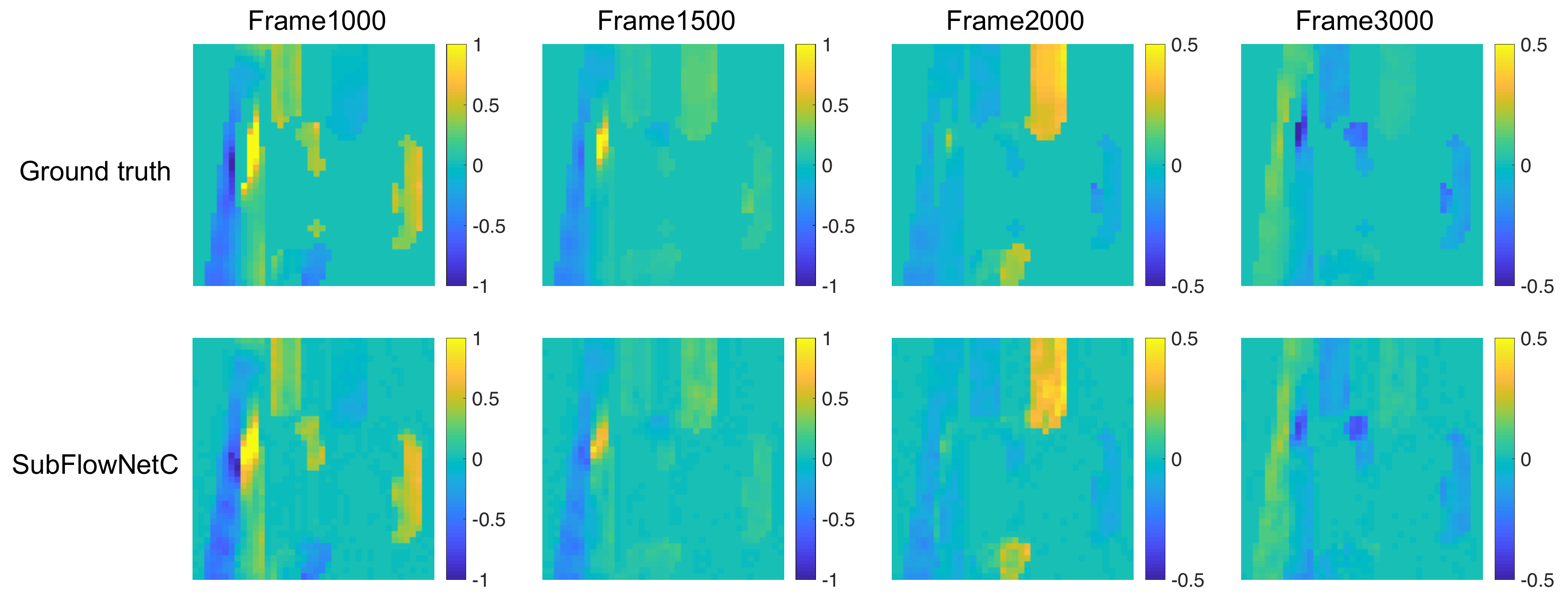}
	\caption{Predicted horizontal full-field displacement of the source testing video frame as shown in Fig. \ref{fig:testing_videos_source}(a)}
	\label{fig:testing_motion_field_source_1_horizontal}
\end{figure}

\begin{figure}[t!]
	\centering\includegraphics[width=1.0\linewidth]{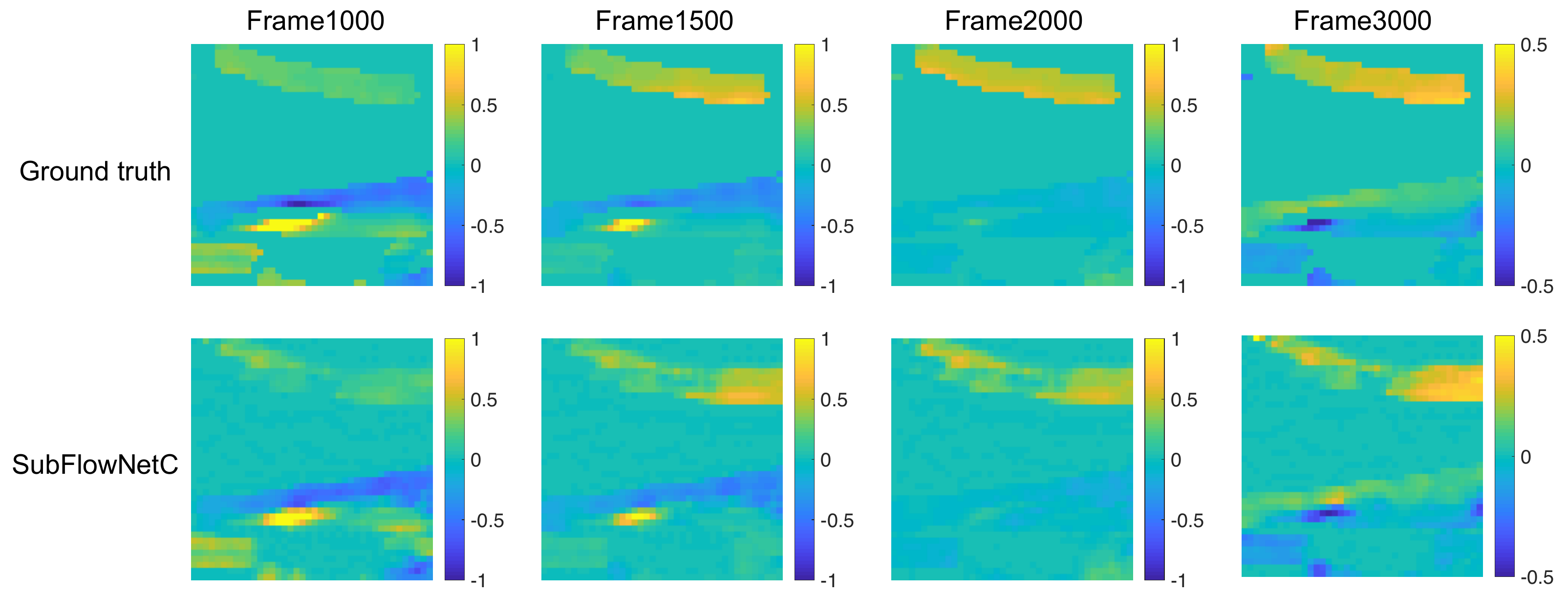}
	\caption{Predicted horizontal full-field displacement of the source testing video frame as shown in Fig. \ref{fig:testing_videos_source}(b)}
	\label{fig:testing_motion_field_source_2_vertical}
\end{figure}

The full displacement field is given to show the performance of the trained network on predicting all pixel motions. Fig. \ref{fig:testing_motion_field_source_1_horizontal} shows the ground truth and predicted full-field horizontal displacements for several frames of the testing video depicted in Fig. \ref{fig:testing_videos_source}(a). Since the phase-based approach fails to produce the displacement on edges as training data, the full displacement fields for only internal \(40\times40\) pixels are presented. It can be seen that the predicted displacement field matches very well the ground truth. The trained network can accurately predict the displacement profile and capture the texture masks of the motion targets including the beam, accelerometer and cables. In the areas without masks, the predicted displacements are close to zero (ground truth). It demonstrates that the network can identify the masks of the detected objects in the input frames. The predicted non-zero motion value for each pixel is also close to the ground truth. The comparison between the predicted vertical displacement field and the ground truth is given in Fig. \ref{fig:testing_motion_field_source_2_vertical}, showing a good performance of the trained network as well. 

\begin{figure}[t!]
	\centering\includegraphics[width=0.95\linewidth]{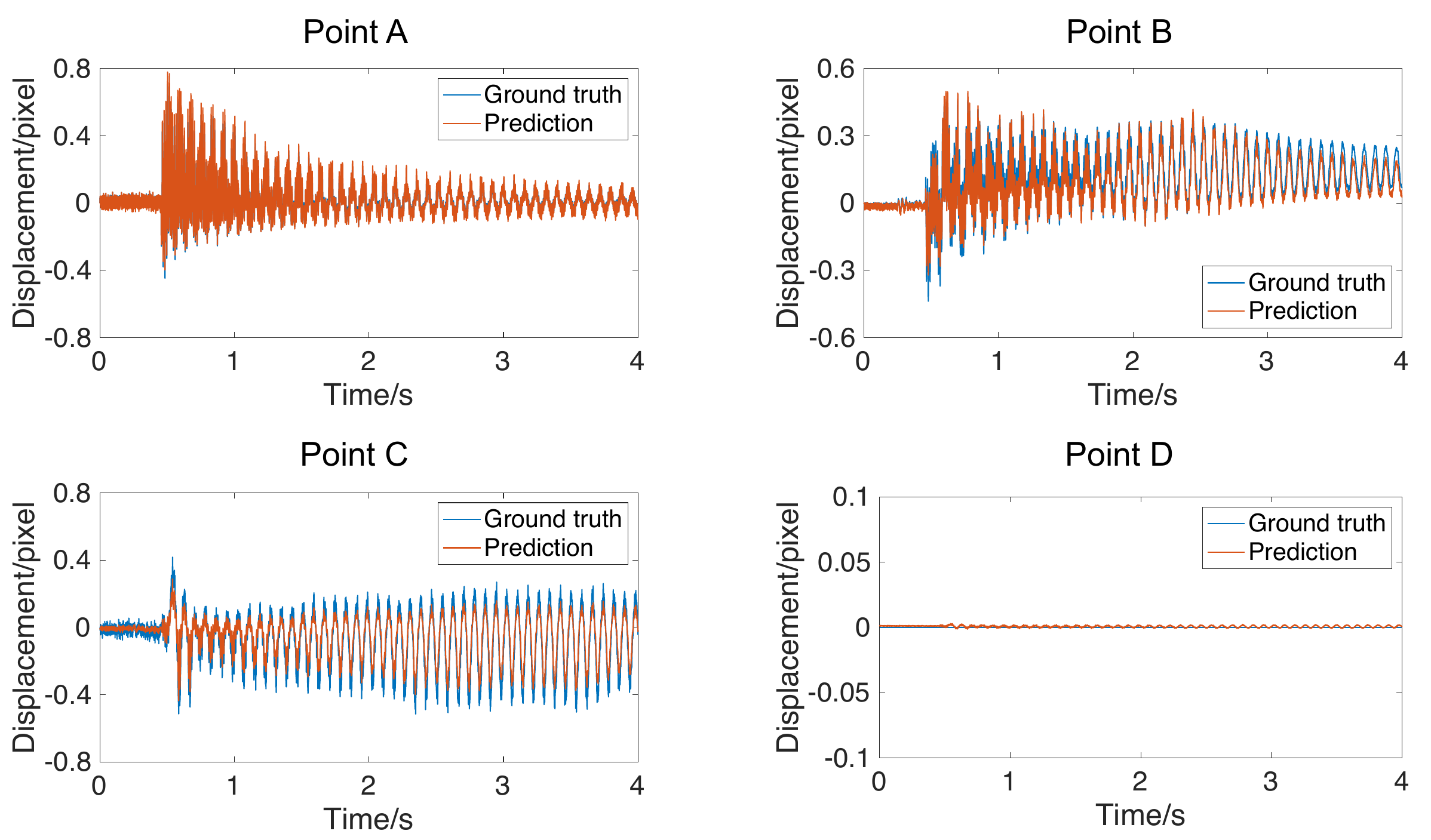}
	\caption{Comparison between the ground truth and the predicted displacement time histories at the annotated points of the testing video shown in Fig. \ref{fig:testing_videos_source}(a). Points A, B and C have texture masks, while point D has no texture mask due to insufficient texture contrast.}
	\label{fig:testing_time_history_source_1}
\end{figure}

\begin{figure}[t!]
	\centering\includegraphics[width=0.95\linewidth]{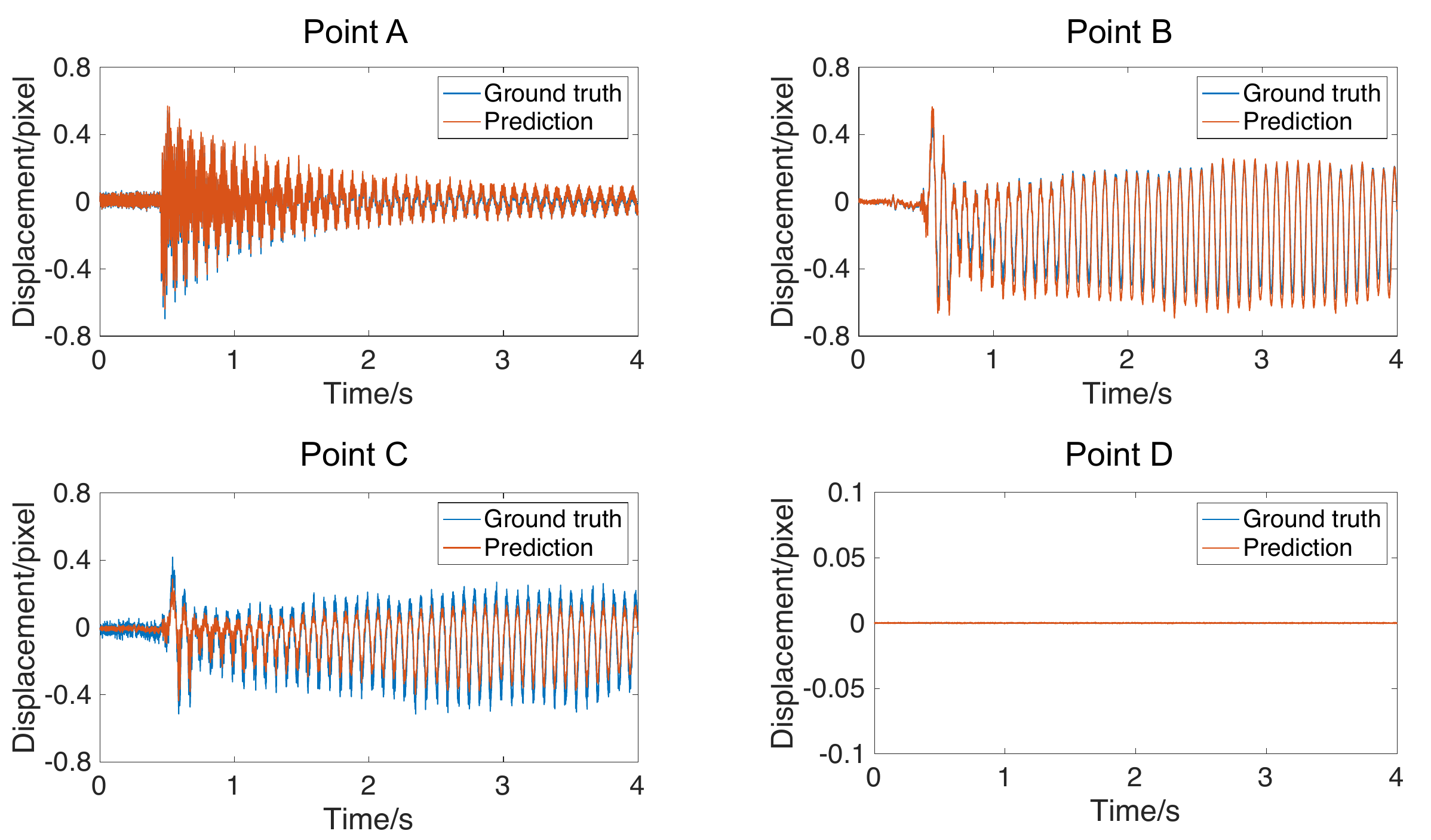}
	\caption{Comparison between the ground truth and the predicted displacement time histories at the annotated points of the testing video shown in Fig. \ref{fig:testing_videos_source}(b). Point A, B and C have texture masks, while point D has no motion target.}
	\label{fig:testing_time_history_source_2}
\end{figure}

The displacement time histories of several annotated points (see Fig. \ref{fig:testing_videos_source}) are presented to further test the performance of the trained network. Those points represent the beam, accelerometer, cables and the area without texture masks. The comparison between the predicted displacement time histories and the ground truth, for the testing videos shown in Fig. \ref{fig:testing_videos_source}(a) and (b), is given in Fig. \ref{fig:testing_time_history_source_1} and Fig. \ref{fig:testing_time_history_source_2} respectively. For the points with texture masks  (points A-C), the trained network can reasonably well predicts the displacement time histories, although minor amplitude errors are in presence (e.g., for point C). For the points without texture masks (point C), the predicted displacement is zero (or extremely close to zero), which illustrates that the trained network can correctly learn and identify the underlying mask information. In particular, the predicted displacement time histories for points A and B are very close to the ground truth (in both amplitude and phase), while the predicted displacements for point C have some deviations compared with the ground truth. The prediction error maybe induced by the limited diversity of dataset, given the fact that only one recorded video is used to generate the training dataset. This issue can be potentially resolved by increasing the diversity of the training data (e.g., based on multiple videos with different objects/targets).

\subsection{Performance on other videos}

After testing on the source video, the generalizability of the trained network is investigated based on videos coming from other lab experiments (e.g., the vibration objects are different). Here, the videos recording the vibration of an aluminum cantilever beam and a three-story building structure \cite{yang2017blind1} are chosen for the validation study.

\begin{figure}[t!]
	\centering\includegraphics[width=0.5\linewidth]{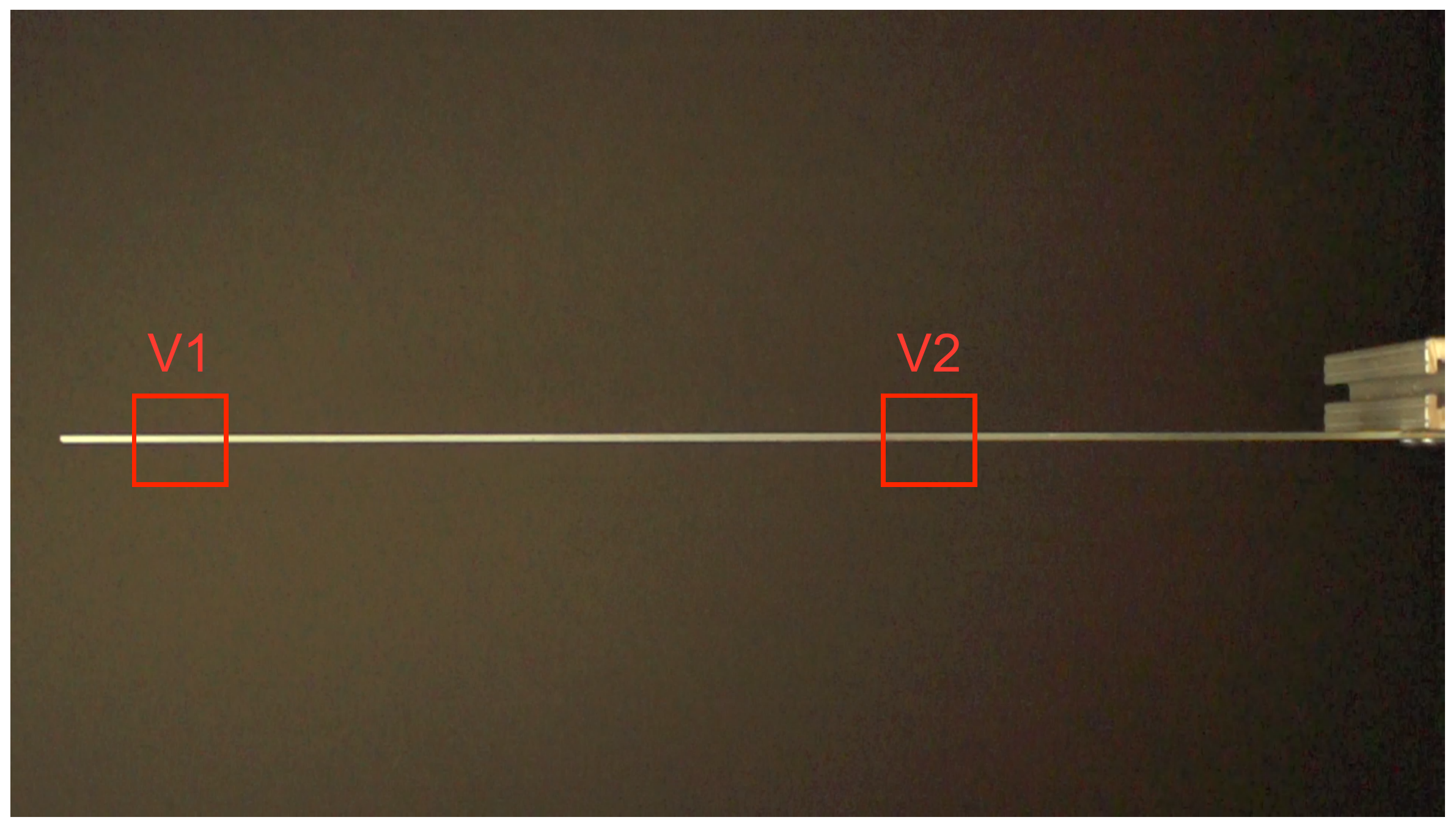}
	\caption{Cantilever beam for testing the trained network for extracting vertical displacements. Two sub-videos (V1 and v2) are cropped from the original video for testing.}
	\label{fig:testing_cantilever_beam}
\end{figure}

\begin{figure}[t!]
	\centering\includegraphics[width=1.0\linewidth]{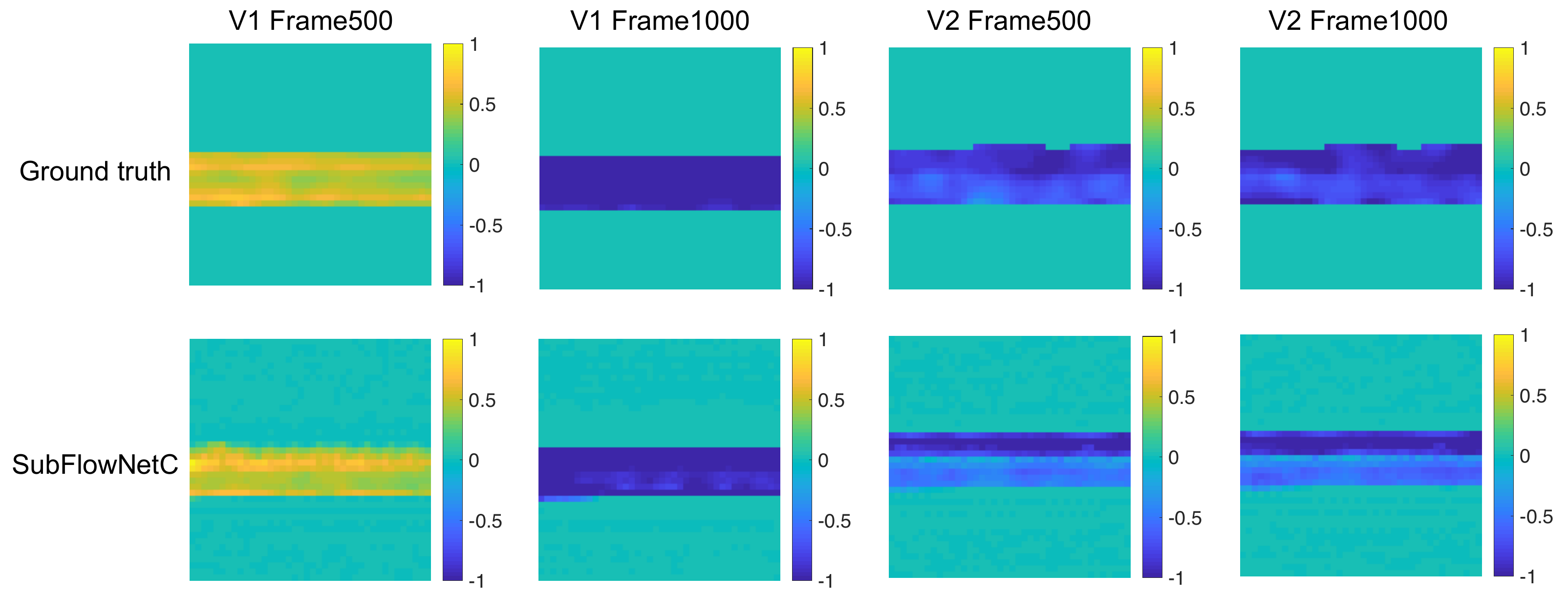}
	\caption{Extracted full-field displacements of the cantilever beam.}
	\label{fig:testing_motion_field_cantilever_beam}
\end{figure}

\begin{figure}[t!]
	\centering\includegraphics[width=0.9\linewidth]{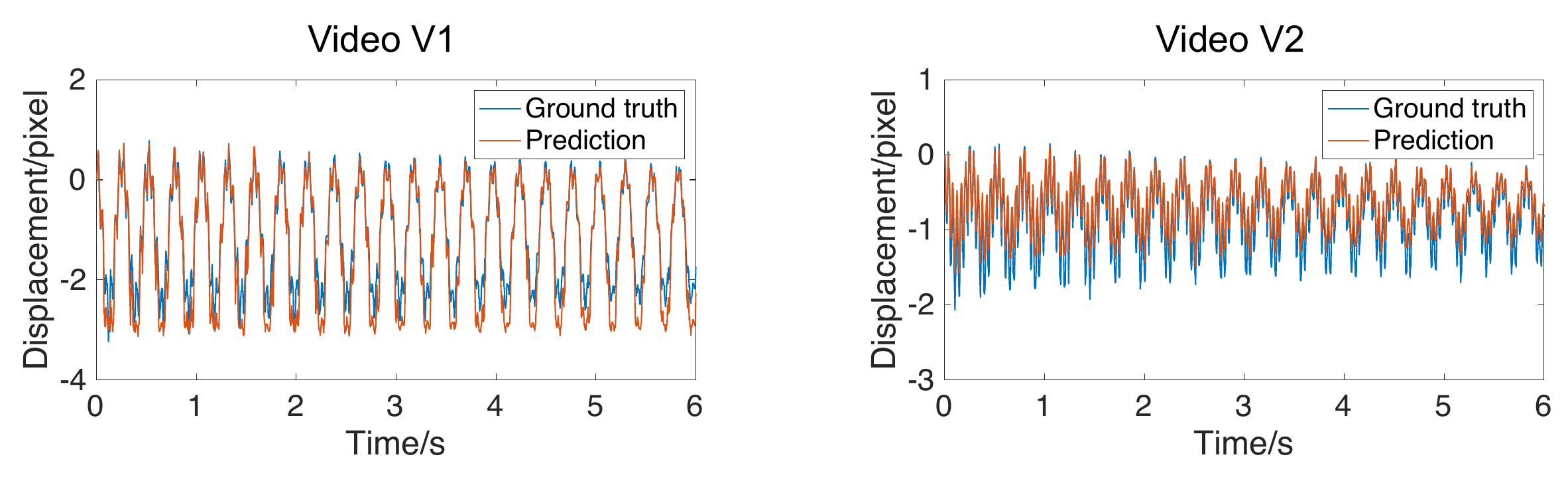}
	\caption{Predicted displacement time histories of the cantilever beam in comparison with the reference ground truth. (a) and (b) represent the displacements of selected pixels with the biggest local amplitudes on the edge of sub-videos v1 and v2 shown in Fig. \ref{fig:testing_time_history_cantilever_beam}.}
	\label{fig:testing_time_history_cantilever_beam}
\end{figure}

\subsubsection{Cantilever beam}\label{cant_beam}

Firstly, the video of vibration of a light cantilever beam shown in Fig. \ref{fig:testing_cantilever_beam} is used to test the performance of the trained network for extracting the full-field vertical displacements. The resolution of this video is \(1,920\times1,080\) and the frame rate is 480 per second. Two sub-videos with size of \(96\times96\) are cropped from the original video and downsampled to \(48\times48\) for testing. Fig. \ref{fig:testing_motion_field_cantilever_beam} shows the predicted displacement fields of frames 500 and 1,000 in two testing videos in comparison with the reference ground truth obtained by the phase-based approach. It can be observed that the trained network successfully captures the beam area and accurately predicts the motions closely to the ground truth. Theoretically, the motions of pixels along the cross-section of the beam should be the same; however, the errors may be induced by the video noise and texture variation of the beam. The predicted displacement time histories of two selected pixels, with the biggest local amplitudes on the edge of the sub-videos, are shown in Fig. \ref{fig:testing_time_history_cantilever_beam}, in comparison with the reference ground truth. It is seen that the predicted displacement is generally close to the ground truth (e.g., accurate phase agreement), with minor amplitude discrepancy.

\subsubsection{Three-story building structure}

\begin{figure}[t!]
	\centering\includegraphics[width=0.7\linewidth]{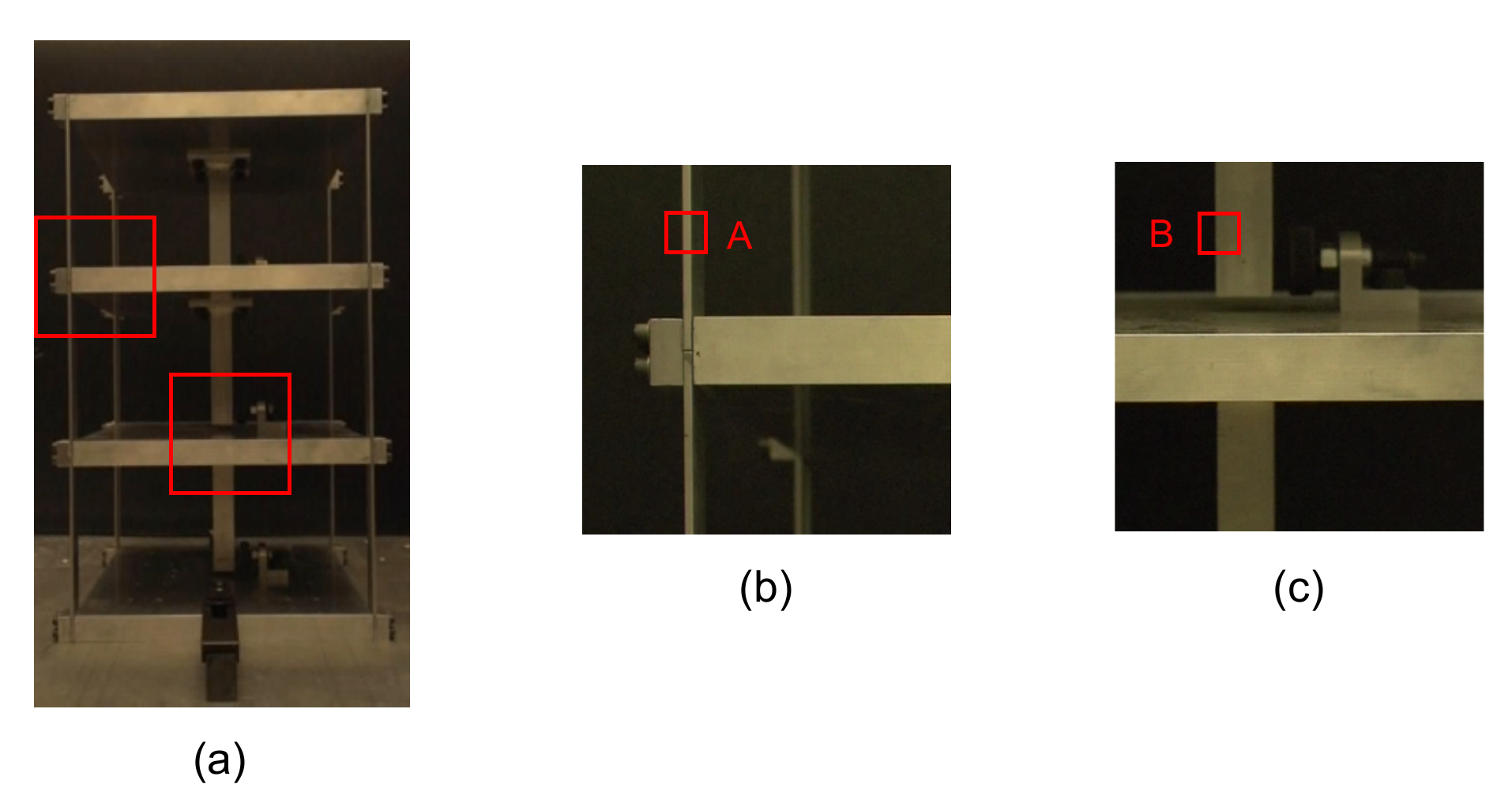}
	\caption{The recorded video of the vibration of a three-story building structure excited by an impact hammer. (a) shows the first frame of the original video. (b) and (c) show the cropped sub-videos from the original video for testing. The annotated points, A and B, show the positions of the studied pixels for displacement time history extraction.}
	\label{fig:testing_frame}
\end{figure}

\begin{figure}[t!]
	\centering\includegraphics[width=0.95\linewidth]{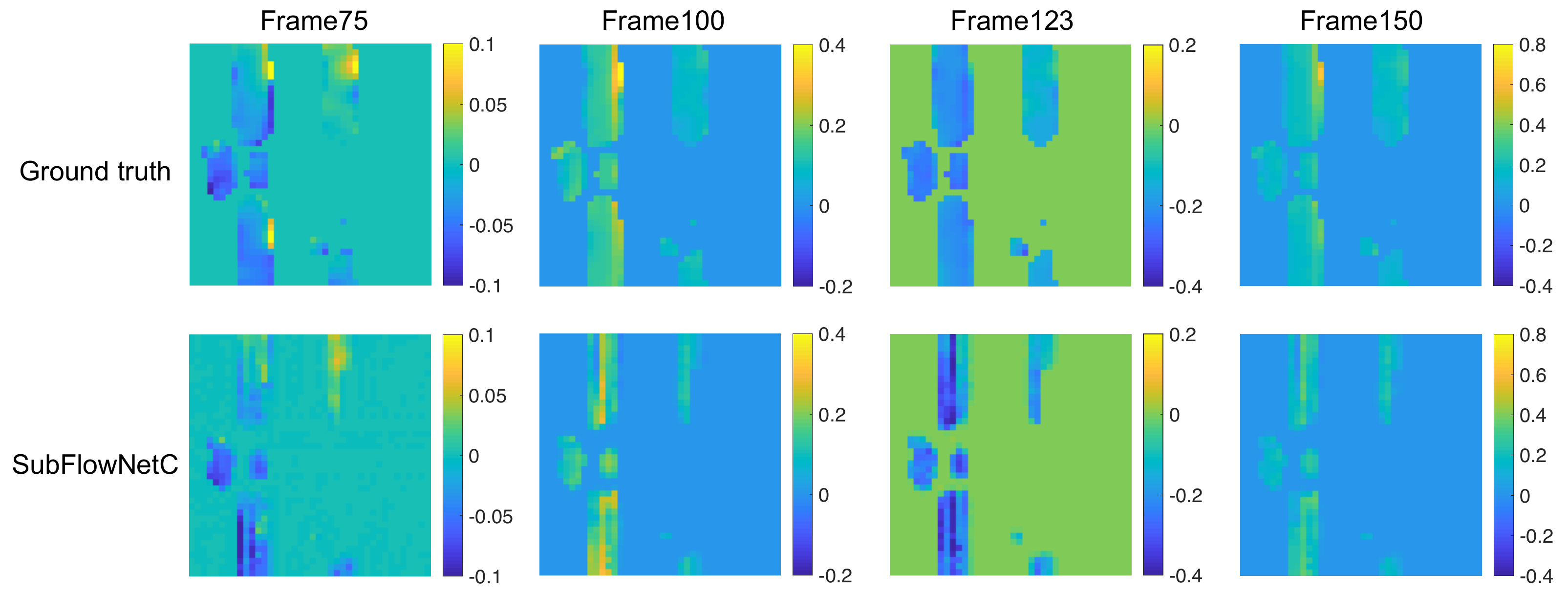}
	\caption{Extracted full-field displacements of the frame structure shown in Fig. \ref{fig:testing_frame}(b)}
	\label{fig:testing_motion_field_frame_b}
\end{figure}

\begin{figure}[t!]
	\centering\includegraphics[width=0.95\linewidth]{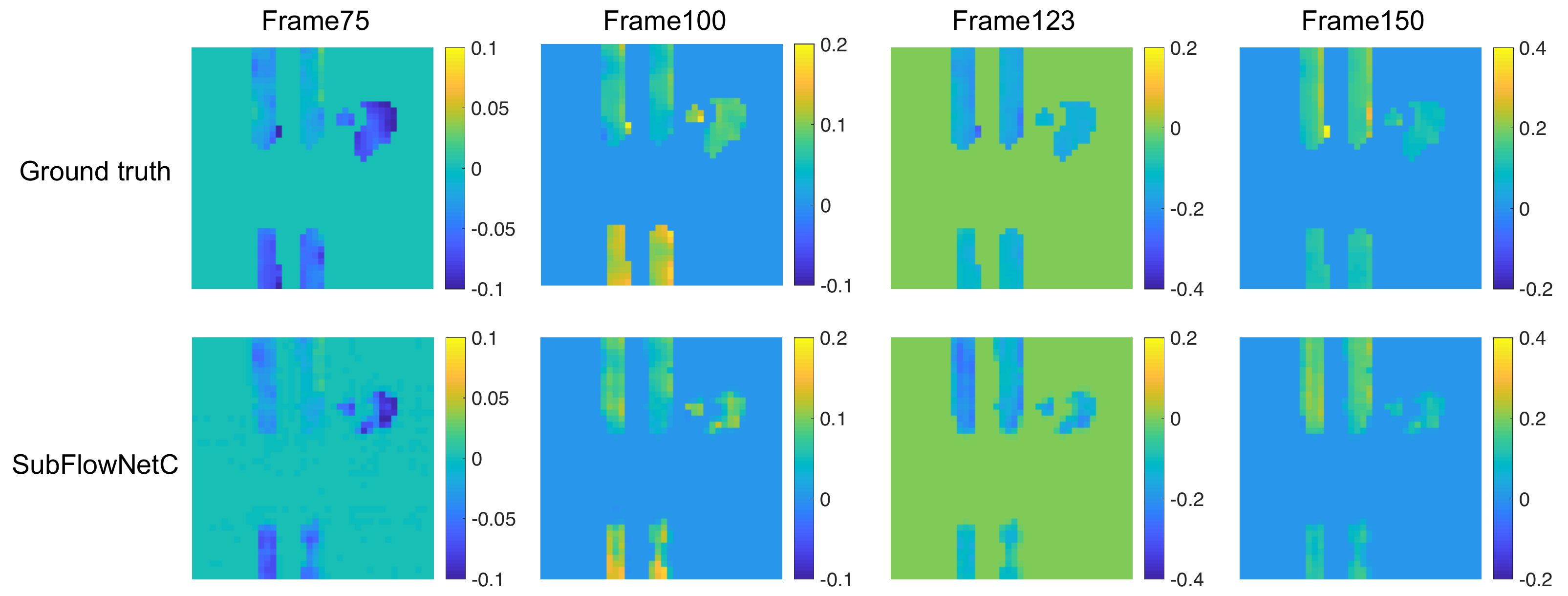}
	\caption{Extracted full-field displacements of the frame structure shown in Fig. \ref{fig:testing_frame}(c)}
	\label{fig:testing_motion_field_frame_c}
\end{figure}

\begin{figure}[t!]
	\centering\includegraphics[width=0.8\linewidth]{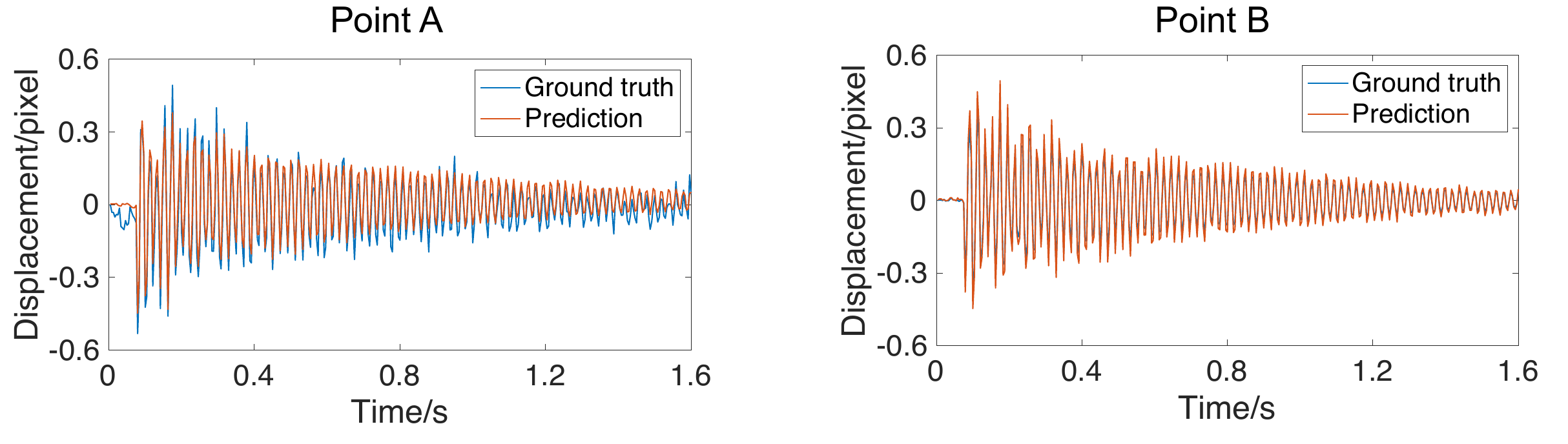}
	\caption{Comparison between the predicted displacement time histories and the ground truth for the annotated points. Points A and B are marked in the testing video shown in Fig. \ref{fig:testing_frame}(b) and (c), respectively.}
	\label{fig:testing_time_history_frame}
\end{figure}

The video of the vibration of a three-story building structure shown in Fig. \ref{fig:testing_frame}(a) is used to further verify the generalizability of the trained network. The  structure was excited by an impact hammer. The resolution of this video is \(1,920\times1,080\) and the frame rate is 240 per second. Fig. \ref{fig:testing_frame} shows the first frame of the video and the cropped sub-videos downsampled to the resolution of 48\(\times\)48. Likewise, both the displacement field and time histories (in the horizontal direction) are extracted. Figs. \ref{fig:testing_motion_field_frame_b} and \ref{fig:testing_motion_field_frame_c} show the predicted displacement field in comparison with the reference ground truth obtained by the phase-based approach. The displacement field can be accurately predicted by the trained network for pixels with clear texture contrast (e.g., the columns shown in Fig. \ref{fig:testing_frame}). Fig. \ref{fig:testing_time_history_frame} shows the extracted displacement time histories of points A and B (see Fig. \ref{fig:testing_frame}) compared with the reference ground truth, which well agree with each other. This illustrative case, as well as the cantilever beam example in Section \ref{cant_beam}, demonstrates that the trained SubFlowNetC network is transferable and generalizable to extraction of full-field displacements from other videos.

\begin{figure}[t!]
	\centering\includegraphics[width=1.0\linewidth]{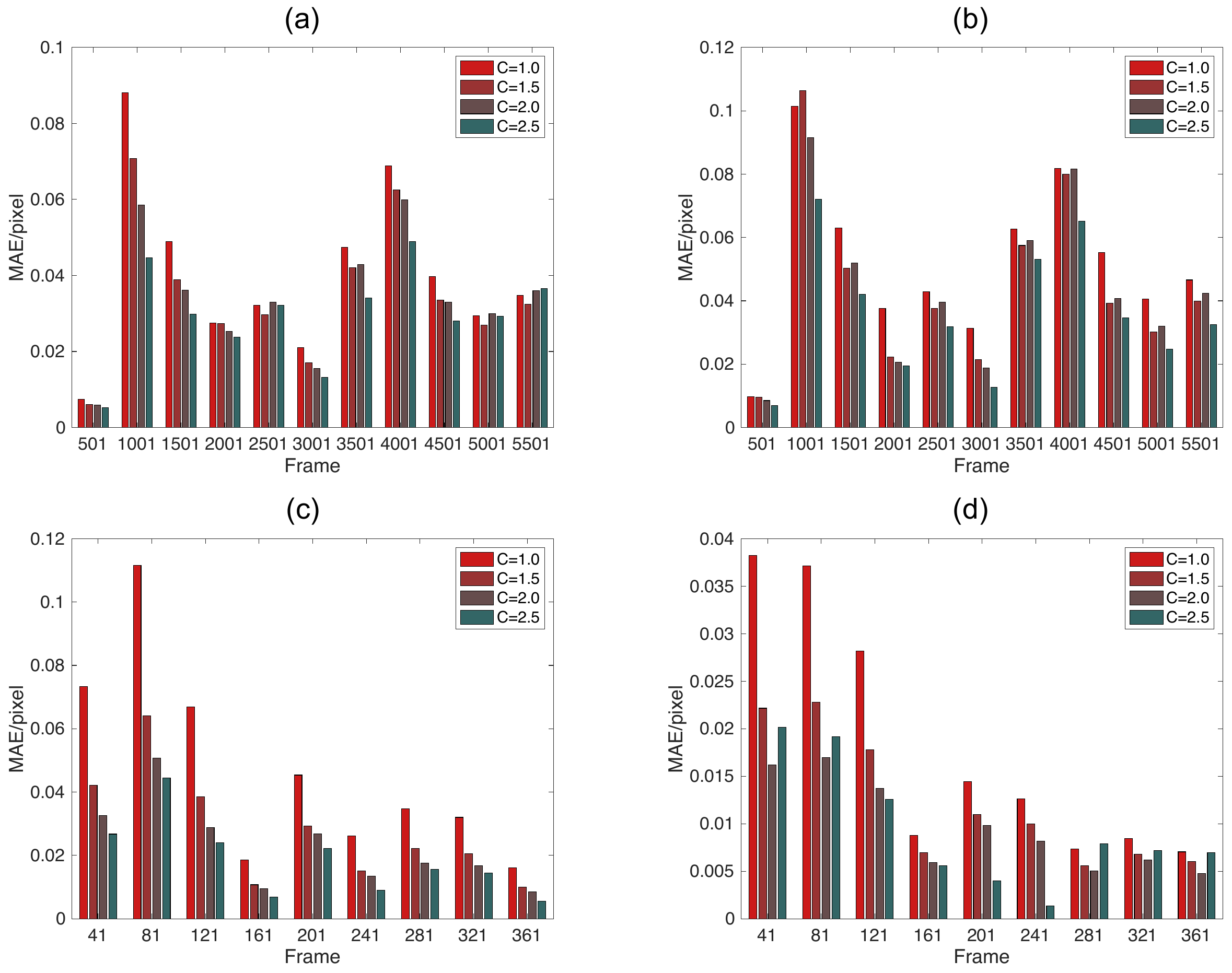}
	\caption{Prediction MAE distribution by SubFlowNetC for some typical frames in the testing videos with different threshold values for the texture mask. (a) and (b) represent the MAE for the source video. (c) and (d) represent the building structure vibration video.}
	\label{fig:texture_accuracy}
\end{figure}

\subsection{Prediction accuracy and pixel contrast}

In the phase-based approach, the local amplitude represents the pixel texture contrast. For full-field dynamic displacement prediction, even the texture mask is considered in the network, the trained network shows different performance on different areas and annotated points with varied amplitude values. Here, the relationship between the local pixel amplitude and the prediction accuracy is analyzed. The index, mean absolute error (MAE) defined as follows, is used to evaluate the prediction accuracy in a frame:
\begin{equation}
\label{MAE}
MAE=\frac{\sum_{i=1}^{N}\left|u_i^R-u_i^P\right|}{N}
\end{equation}
where \(N\) denotes the number of pixels in a frame; \(u_i^R\) and \(u_i^P\) denote respectively the real and predicted displacement at each pixel. Here, the effect of local amplitude on the prediction accuracy is investigated by varying the threshold value for the texture mask. The initial threshold is chosen 1/5 of the mean of the 30 pixels with the greatest amplitudes. The threshold value varies as follows:
\begin{equation}
\label{varied_threshold}
T=C\cdot\ T_0
\end{equation}
where \(T\) is the varied threshold value for the texture mask; \(T_0\) is the initial threshold value; \(C\) is the coefficient used to change the threshold value. After the new threshold value determined, the ground truth and the predicted displacement fields whose amplitudes are above the threshold are expressed as
\begin{equation}
\label{new_motion_field}
M^T=M_0\cdot\ m^T
\end{equation}
where \(M_0\) is the displacement field for the initial texture threshold value, \(m^T\) is the texture mask with a new threshold, and \(M_T\) is the displacement field accounting for the new texture mask. The prediction error of pixels with the new texture threshold value is calculated by Eq. \eref{MAE}. Fig. \ref{fig:texture_accuracy} shows the MAE of some typical frames of the testing videos (i.e., the source video and the building structure vibration video) with varying threshold values. It is seen that the prediction MAEs for majority of the given frames are less than 0.1 pixel, which means the trained SubFlowNetC network possesses a high accuracy for full field displacement extraction. In general, the MAEs decrease along with the increase of the threshold value, especially for threshold values from 1.0 to 1.5. The trained network has a better prediction performance for pixels with larger local amplitude values.

\subsection{Discussions on learned filters }

\begin{figure}[t!]
	\centering\includegraphics[width=1.0\linewidth]{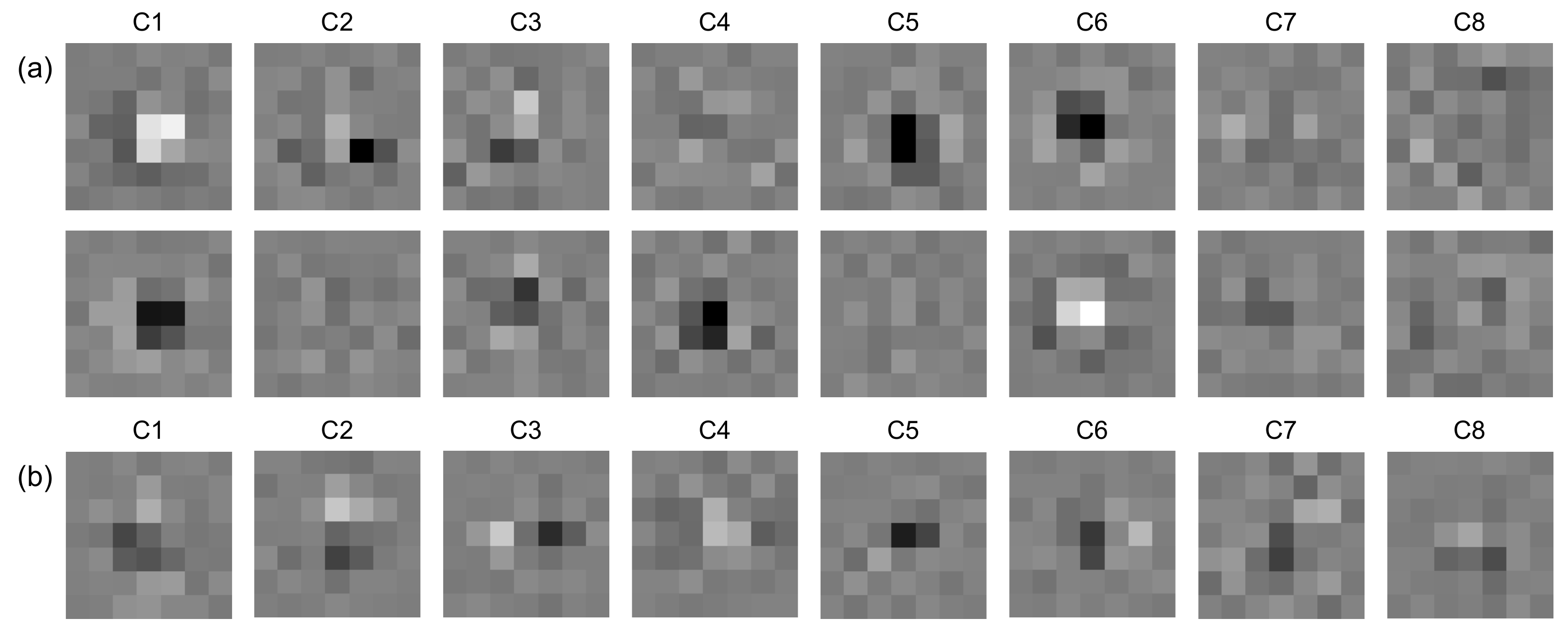}
	\caption{Visualization of the learned filters in the first Conv layer. (a) and (b) respectively show the filters of SubFlowNetS and SubFlowNetC. For both networks, there are 8 kernels in the first Conv layer.}
	\label{fig:learned_filters}
\end{figure}

It was found in \cite{ranjan2017optical} that the learned filters (weights) appear similar to traditional Gaussian derivative filters used by classical optical flow estimation methods for extracting motion representatives. In the phase-based displacement extraction approach, the designed quadrature complex filters process the images to obtain the local phase as the motion representative. Fig. \ref{fig:learned_filters} visualizes the learned filters of the first Conv layer for SubFlowNetS and SubFlowNetC. It is seen that many of the filters are similar to the complex filters depicted in Fig. \ref{fig:quadrature_filters}. These learned filters also appear to be similar to the traditional derivative filters in variational approaches for optical flow estimation \cite{baker2011database,ranjan2017optical}. In addition, the motion of each pixel is related to its surrounding pixels as noted in the learned filters. The filters have bigger values in the core, indicating that the motion of each pixel is affected more by its neighbor pixels.

\section{Conclusions} \label{conclusions}

This paper presents a deep learning approach based on convolutional neural networks to extract the full-field subtle displacement of structural vibration. In particular, two network architectures, SubFlowNetS and SubFlowNetC, are designed in an encoder-decoder scheme. In order to account for the sparsity of the motion field, a texture mask layer is added at the end of each network while the networks are trained with the supervision of both full and sparse motion fields via the stacked loss function. The training dataset is generated from a single lab-recorded high-speed video, where image pairs are taken as input while the phase-based approach is adopted to obtain the full-field subpixel motion field as output labels. The performance of trained networks is demonstrated by extracting the horizontal or vertical motion field from various recorded videos. The results illustrate that the proposed networks, despite trained against limited labeled datasets based on a single video, has the capacity to extract the full-field subtle displacements and possesses generalizability for other videos with different motion targets. With the supervision of both full and sparse motion fields, the trained networks are able to identify the pixels with sufficient texture contrast as well as their displacement time histories. The effect of texture contrast on prediction accuracy of the trained networks is also investigated. Motions of the pixels with larger local amplitude value tend to be easier to captured by the network. Moreover, the learned filters of the convolution layers demonstrate the capacity for nonlinear mapping, showing similarity to the complex filters in the phase-based approach and the traditional derivative filters in variational approaches for optical flow estimation. Given the salient feature discussed above, the trained networks have potential to enable the monitoring of structural vibration in real time. The focus of our future study will be placed on application and validation of this technique on real-world structures.

\section*{Acknowledgement}
\small
This work was supported in part by Federal Railroad Administration under grant FR19RPD3100000022, which is greatly acknowledged. Y. Yang would like to acknowledge the support by the Physics of AI program of Defense Advanced Research Projects Agency (DARPA). The authors thank Dr. Justin G. Chen from MIT Lincoln Laboratory for sharing the recorded video which was used to verify the proposed approach.
\normalsize

\bibliographystyle{elsarticle-num}
\bibliography{refs}

\end{document}